\documentclass[letterpaper,lang=cn]{article} 
\usepackage{aaai25}  
\usepackage{times}  
\usepackage{helvet}  
\usepackage{courier}  
\usepackage[hyphens]{url}  
\usepackage{graphicx} 
\usepackage{natbib}  
\usepackage{caption} 
\usepackage{float}
\usepackage{algorithmic}
\usepackage[linesnumbered, ruled]{algorithm2e}

\usepackage{newfloat}
\usepackage{listings}
\DeclareCaptionStyle{ruled}{labelfont=normalfont,labelsep=colon,strut=off} 
\newcommand\YAMLcolonstyle{\color{red}\mdseries}
\newcommand\YAMLkeystyle{\color{black}\bfseries}
\newcommand\YAMLvaluestyle{\color{blue}\mdseries}

\makeatletter

\expandafter\expandafter\expandafter\lstdefinelanguage
\expandafter{yaml}
{
  keywords={true,false,null,y,n},
  keywordstyle=\color{darkgray}\bfseries,
  basicstyle=\YAMLkeystyle,                                 
  sensitive=false,
  comment=[l]{\#},
  morecomment=[s]{/*}{*/},
  commentstyle=\color{purple}\ttfamily,
  stringstyle=\YAMLvaluestyle\ttfamily,
  moredelim=[l][\color{orange}]{\&},
  moredelim=[l][\color{magenta}]{*},
  moredelim=**[il][\YAMLcolonstyle{:}\YAMLvaluestyle]{:},   
  morestring=[b]',
  morestring=[b]",
  literate =    {---}{{\ProcessThreeDashes}}3
                {>}{{\textcolor{red}\textgreater}}1     
                {|}{{\textcolor{red}\textbar}}1 
                {\ -\ }{{\mdseries\ -\ }}3,
}

\usepackage{xcolor}         

\usepackage{color}
\usepackage{colortbl}
\usepackage{graphicx}

\usepackage{amsmath}
\usepackage{amsfonts}
\usepackage{subfigure}
\usepackage{booktabs}

\title{Modeling All Response Surfaces in One for Conditional Search Spaces}
\author{
    Jiaxing Li\textsuperscript{\rm 1,\rm 2}\thanks{This work was done while the author was an intern at JD Explore Academy.},
    Wei Liu\textsuperscript{\rm 2},
    Chao Xue\textsuperscript{\rm 2},
    Yibing Zhan\textsuperscript{\rm 2}\footnote{Corresponding Authors.},
    Xiaoxing Wang\textsuperscript{\rm 3},
    Weifeng Liu\textsuperscript{\rm 1},
    Dacheng Tao\textsuperscript{\rm 4}
}
\affiliations{
    \textsuperscript{\rm 1}China University of Petroleum (East China) \\
    \textsuperscript{\rm 2}JD Explore Academy \\
    \textsuperscript{\rm 3}Shanghai Jiao Tong University \\
    \textsuperscript{\rm 4}Nanyang Technological University\\
    xinghui0606@gmail.com, lwwd1996@163.com, xuechao19@jd.com, zybjy@mail.ustc.edu.cn, figure1\_wxx@sjtu.edu.cn, liuwf@upc.edu.cn, dacheng.tao@gmail.com
}

\begin{document}

\maketitle


\begin{abstract}
Bayesian Optimization (BO) is a sample-efficient black-box optimizer commonly used in search spaces where hyperparameters are independent. However, in many practical AutoML scenarios, there will be dependencies among hyperparameters, forming a conditional search space, which can be partitioned into structurally distinct subspaces. 
The structure and dimensionality of hyperparameter configurations vary across these subspaces, challenging the application of BO. Some previous BO works have proposed solutions to develop multiple Gaussian Process models in these subspaces. However, these approaches tend to be inefficient as they require a substantial number of observations to guarantee each GP's performance and cannot capture relationships between hyperparameters across different subspaces.
To address these issues, this paper proposes a novel approach to model the response surfaces of all subspaces in one, which can model the relationships between hyperparameters elegantly via a self-attention mechanism. Concretely, we design a structure-aware hyperparameter embedding to preserve the structural information. Then, we introduce an attention-based deep feature extractor, capable of projecting configurations with different structures from various subspaces into a unified feature space, where the response surfaces can be formulated using a single standard Gaussian Process.
The empirical results on a simulation function, various real-world tasks, and HPO-B benchmark demonstrate that our proposed approach improves the efficacy and efficiency of BO within conditional search spaces.
\end{abstract}

\section{Introduction}
Bayesian Optimization (BO)~\cite{review,Garnett_2023,Baxus,Bounce} is a powerful and efficient global optimizer for expensive black-box functions, which has gained increasing attention in AutoML systems and achieved great success in a number of practical application fields in recent years~\cite{DBLP:robotic_Martinez-CantinFDC07,bo_bio_genedesign,DBLP:robotic_journals/amai/CalandraSPD16,roussel2024bayesian}. Considering a black-box function $f$: $\chi \to \mathbb{R}$ defined on a search space $\chi$, BO aims to find the global optimal configuration
$x_{*}=\mathop{\arg\min}_{x\in{\chi}} {f(x)}$.
The sequential Bayesian optimization procedure contains two key steps~\cite{review}: (1) BO seeks a probabilistic surrogate model to capture the distribution of the black-box $f$ given $n$ noisy observations $y_{i} = f(x_{i}) + \epsilon, i \subset {1,...,n}, \epsilon \sim \mathcal{N}(0, \sigma)$. (2) Suggest the next query $x_{n+1}$ by maximizing an exploit-explore trade-off acquisition function $\alpha(x)$. The most common choice of the surrogate models is Gaussian Process (GP)~\cite{DBLP:spearmint_conf/nips/SnoekLA12,DBLP:journals/ijns/Seeger04} due to its generality and good uncertainty estimation. As to acquisition functions, the common choice for GP-based BO is the Expected Improvement (EI)~\cite{DBLP:ei_journals/jgo/Mockus94,UI2EI}.

In the traditional BO setting, the search space $\chi$ is flat where all configurations have the same dimensions and structure~\footnote{The structure of a configuration in this paper is defined as containing two aspects: dependencies between pairs of hyperparameters and semantic information of each hyperparameter. Therefore, the hyperparameter ``learning rate'' has the same semantic information in XGBoost and DNN models.}: $x\in{\chi} \subset \mathbb{R}^d$, where $d$ is the number of dimensions. However, in many practical AutoML scenarios, such as Combined Algorithm Selection and Hyperparameter optimization (CASH)~\cite{arc} and Neural Architecture Search (NAS)~\cite{DBLP:conf/kdd/ThorntonHHL13,MNAS}, the search spaces are conditional where the configurations have different structures and number of dimension.

\begin{figure*}[ht]
    \centering       \includegraphics[width=1.6\columnwidth]{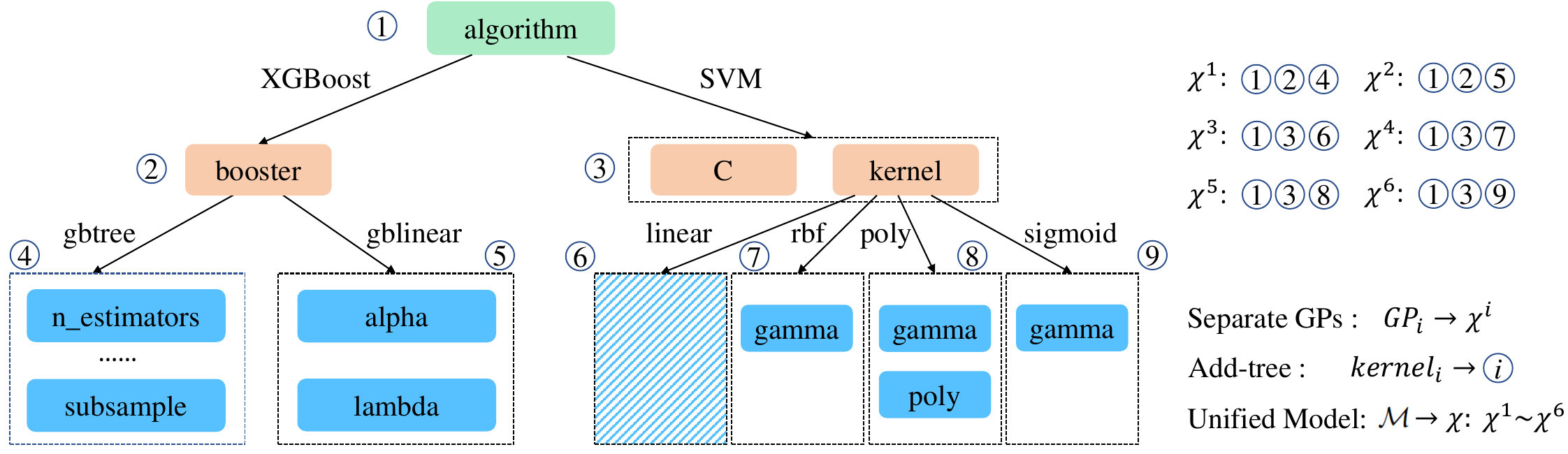}  
\caption{An example of the tree-structured search space for a CASH task. The space contains two popular algorithms and their distinct hyperparameters. According to the dependencies among hyperparameters, the search space can be formed as 9 nodes and partitioned into 6 flat subspaces. 
The approaches using separate GPs build a model $GP_i$ for each subspace and conduct optimization in each subspace. Add-tree~\cite{addTree_tpami} builds a kernel on each node and integrates them using the additive assumption. Our method explores to build a unified surrogate model $\mathcal{M}$ for all subspaces $\chi^1 \sim \chi^6$.}
\label{fig:new_tree_sp}  
\end{figure*}

Such a conditional search space $\chi$ can be decomposed into a series of flat subspaces~\cite{Tree-Structured}: $\chi =\chi^{1} \cup \chi^{2} \cup ... \cup \chi^{n}$ and configurations in the same subspace have the same structure: $x^{i} \in \chi^{i} \subset \mathbb{R}^{d^{i}}$. We give a typical CASH example in Fig.~\ref{fig:new_tree_sp}. The space consists of two machine-learning algorithms and their specific hyperparameters, which can be partitioned into 6 subspaces.
Due to the unaligned features among configurations within different subspaces, the surrogate models commonly used, such as GP, are not directly applicable. 

A straightforward strategy to solve this issue is to build separate surrogate models for each subspace independently~\cite{DBLP:conf/ijcnn/LevesqueDGS17}.  Given the lack of information sharing during the training phase of the GPs, most of these approaches suffer from the requirement of a significant number of observations to ensure the predictive performance of each GP~\cite{addtree_icml}. Recent work~\cite{addTree_tpami} proposed to set a covariance function on each node of the tree-structured space and integrate them in a GP through an additive assumption. However, the model is still essentially divided, with the parameters not shared among different nodes. As a consequence, the relationships between non-shared but semantically related hyperparameters, such as "gamma" of different kernels (in nodes 7-9 of Fig.~\ref{fig:new_tree_sp}), are totally ignored.

To overcome these limitations, we explore building a unified surrogate model for all hyperparameters within the tree-structured search space, by which we can capture the relationships among all hyperparameters and utilize the information of all observations even from different subspaces.
Concretely, to obtain a better representation of the hyperparameter's features, we propose a hyperparameter embedding that preserves the structural information of the subspace each hyperparameter belongs to. In our framework, we treat the embedding of each hyperparameter as a token and each hyperparameter configuration as a sequence of tokens. Then, we introduce an attention-based deep encoder, which is capable of modeling the global relationships among tokens using self-attention blocks and projecting the sequences from different subspaces into a unified latent space by an average pooling operator. With the attention-based encoder, the features of these configurations become comparable and can be modeled by any standard kernel function. 
Different from the previous works, our approach considers the relationships among hyperparameters in different subspaces, and the parameters of the surrogate model are shared and completely consistent for all observations, which can improve the sample efficiency of BO in the tree-structured search spaces. 

Some recent works have proposed using variational autoencoders (VAEs)~\cite{vae} to transform high-dimensional and structured inputs into continuous, low-dimensional spaces that are more amenable to Bayesian optimization techniques~\cite{GVAE, SVAE-BO, W-LBO, T-LBO, LOL-BO}. Structured or combinatorial inputs~\cite{DBLP:conf/nips/DeshwalD21} refer to sequences, trees, or graphs organized by categorical variables. In contrast, the conditional search space we address has no such restrictions on variable types involving both numerical and categorical variables.
Consequently, these existing methods do not readily extend to search spaces containing both categorical and numerical hyperparameters in a complex, structured relationship, and we don't consider these methods as our competitors.

In conclusion, our contributions can be summarized as follows:

1) We explore the relationships among all hyperparameters and integrate information from all observations with higher sample efficiency via a unified surrogate model.

2) We propose a novel attention-based Bayesian optimization framework, consisting of two key components: a hyperparameter embedding to preserve the structural feature of each hyperparameter in a configuration and an attention-based deep kernel Gaussian process to model the response surfaces of all subspaces in one. In our framework, all parameters are shared for any configurations, making it consider the relationships among all hyperparameters.

3) We conduct experiments on a standard tree-structured simulation function, a Neural Architecture Search (NAS) task, and several real-world OpenML tasks to demonstrate the efficiency and efficacy of our proposed approach. Besides, to validate the warm-starting capability of our method in scenarios with extensive historical data, we also conduct a meta-learning experiment on the HPO-B benchmark.

\section{BO for Conditional Search Space}
Sequential Model-based Algorithm Configuration (SMAC) is an early BO method that can deal with conditional search spaces~\cite{SMAC}. SMAC involves utilizing all potentially active hyperparameters across the entire space as inputs to the surrogate model. Before being fed into the surrogate model, inactive hyperparameters in configurations from different subspaces will be filled with default values to minimize their interference with the surrogate model. This approach will introduce redundant codes, resulting in higher-dimensional problems and diminishing the efficiency of fitting the surrogate model.

Compared to SMAC, GP-based BO gives better uncertainty estimation and shows higher sample efficiency in practical applications. The most straightforward way to leverage GP in the context of conditional search spaces is to build separate GPs in these subspaces~\cite{DBLP:conf/ijcnn/LevesqueDGS17, bandits-bo}. Due to the lack of an information-sharing mechanism during the training stage, these approaches require a large number of observations in each subspace to fit these models, making it impractical when the number of subspaces becomes too large~\cite{Tree-Structured,addtree_icml}.
\citet{Tree-Structured} proposed a semi-parametric GP method that captures the relationship of GPs via a weight vector. However, the linear relationship assumption among these GPs makes it less flexible. Following this work, ~\citet{addtree_icml} proposed an Add-Tree covariance function to capture the response surfaces. It sets a covariance function on each node of the tree-structured space and integrates them using the additive assumption. However, it totally ignores the relationships between non-shared hyperparameters.

\section{Preliminaries}
\paragraph{Deep Kernel Learning for Gaussian Process}
The standard GPs rely on a suitable handcrafted kernel function. An inappropriate kernel will lead to sub-optimal performances due to the false assumptions~\cite{HEBO}. The idea of deep kernel learning~\cite{DKL} is to introduce a neural network $\phi$ to transform the configuration $x$ to a latent representation that serves as the input of the kernel, which facilitates learning the kernel in a suitable space. Concretely, the kernel function is shown as:
\begin{align} 
\label{eq:DeepKernel}
k_{deep}(x, x^{'} | \theta, \omega) = k(\phi(x, \omega), \phi(x^{'}, \omega) | \theta),
\end{align}
where $\omega$ represents the weights of the deep neural network $\phi$ and $\theta$ represents the parameters of the kernel function. All these parameters can be jointly estimated by maximizing the marginal likelihood~\cite{FSBO}:
\begin{align}
\label{loglikelihood}
\log p(\mathbf{y|X}, \theta, \omega) 
& \propto - (\mathbf{y}^{T} \mathbf{K}_{deep}^{-1} \mathbf{y} + \log(|\mathbf{K}_{deep}|)),
\end{align}
where $\mathbf{X}$ and $\mathbf{y}$ represent the configurations and their noisy response, respectively, and the deep kernel matrix $\mathbf{K}_{deep}$ is equal to $k_{deep}(\mathbf{X}, \mathbf{X} | \theta, \omega) + \sigma^{2} \mathbf{I}$~\cite{DBLP:journals/ijns/Seeger04}.

\paragraph{Self-Attention Mechanism}
In the field of Natural Language Processing (NLP), the transformer model~\cite{DBLP:conf/nips/VaswaniSPUJGKP17} is a pioneering work, which utilizes the self-attention mechanism to model the global relationship between different words in a sequence, such as a sentence and paragraphs. In many later practices and papers, the effectiveness of the attention module was verified and applied in many fields~\cite{gps-net,DBLP:journals/aiopen/LinWLQ22}. For an input sequence of $N$ words, the $d_k$-dimensional embeddings of words plus the corresponding positional embeddings are fed into a stacked attention module. In the attention mechanism, the packed matrix representation of the query $Q \in \mathbb{R}^{N \times d_k}$, the key $K \in \mathbb{R}^{N \times d_k}$ and the value $V \in \mathbb{R}^{N \times d_k}$ are fused through $\operatorname{Attention}(\mathbf{Q}, \mathbf{K}, \mathbf{V})=\operatorname{softmax}\left(\frac{\mathbf{Q K ^ { \top }}}{\sqrt{d_{k}}}\right) \mathbf{V}=\mathbf{A V}$. The attention matrix $A$ contains the similarity between each pair of words, which makes the output feature of a word a fusion of the feature of each word in the whole sequence. In this paper, this mechanism is employed to model the relationship among hyperparameters in a tree-structured search space, where the sampled configuration can be viewed as a sequence of hyperparameters.

\section{Methodology}
In this section, we present an attention-based Bayesian optimization framework (AttnBO) to model the relationships among all hyperparameters and build a unified response surface for all subspaces. 
First, we give the modeling of the conditional and tree-structured search space, and we analyze the differences between the unified surrogate model and multiple independent surrogate models. Then we provide the methodological details for each component of our approach.
\subsection{Problem Formulation}
\label{Problem Formulation}

A conditional search space $\chi$ can be decomposed into a series of flat subspaces~\cite{Tree-Structured,addtree_icml}: $\chi =\chi^{1} \cup \chi^{2} \cup ... \cup \chi^{n}$ and configurations in the same subspace have the same structure: $x^{i} \in \chi^{i} \subset \mathbb{R}^{d^{i}}$, $d^{i}$ means the number of dimensions of the subspace $\chi^{i}$. 

Here, we give a loose assumption that such a conditional structure is a tree structure or a combination of several tree structures. To avoid multiple repetitions and simplify the notation, we use a single tree structure that refers to multiple trees since the embedding method and our method can handle both cases in a unified way. Inspired by~\citet{addtree_icml}, we define the search space $\chi$ as a tree structure $\mathcal{T}=(V, E)$, where one node $v \in V$ refers to one set of hyperparameters, each with the associated range, type and value (if sampled), and $e \in E$ refers to the dependency relationship between a node $v$ and the father node $v\uparrow \in V$, which is caused by the hyperparameter $p \in v$ and the dependent hyperparameter $p\uparrow  \in v\uparrow$. We assign a virtual father vertex with a fixed embedding for those root nodes, which can also be applied to the case of multiple trees. The ancestor nodes and intermediate nodes have at least one decision variable.  
Each subspace $\chi^{i}$ is a path from the root to a leaf. 

After sampling from the search space $\chi$, we group the configurations $\mathbf{X}$ and their noisy responses by each subspace $\chi^{i}$ as $\mathbf{X}^i = \left\{x^{i}_j\right\}$, $\mathbf{y}^i = \left\{y^{i}_j\right\}$ where $i = 1, 2, ..., n$ and $j = 1, 2, ..., N^{i}$, $N^{i}$ represents the number of sampled points belonging to the subspace $\chi^{i}$. We denote the observations as $D = \left\{D^{1}, D^{2}, ..., D^{n}\right\}$, where $D^{i}$ means all observations $\left\{(x_{j}^{i}, y_{j}^{i})| j = 1,2,...,N^{i}\right\}$ in subspace $\chi^{i}$. A configuration $x^{i}_{j}$ is a sequence of hyperparameters $ \left [p_k^{x^{i}_j} \right ]$ with specific values, where $k=1, 2, ..., d^{i}$, $d^{i}$ means the dimension (the number of hyperparameters) of the subspace $\chi^{i}$. 

Most previous works~\cite{Tree-Structured,bandits-bo,DBLP:conf/ijcnn/LevesqueDGS17} proposed to build a surrogate model $M^i$ for each subspace $\chi^{i}$ independently. These surrogate models can only be trained on the observations $D^{i}$ and predict the posterior distribution $P(f_{*}^{i}|\mathbf{X}^{i}, \mathbf{y}^i, x_{*}^{i})$, where $x_{*}^{i}$ represents a configuration from subspace $\chi^{i}$ and $f_{*}^{i}$ represents the objective funtion value on $x_{*}^{i}$. In contrast, a unified surrogate model $M$ can be trained on all observations $D$ and infer posterior probabilities for configuration $x_{*}$ within any subspace: $P(f_{*}|\mathbf{X}, \mathbf{y}, x_{*})$, which implies it must handle the input configurations that vary in both dimension and semantics while ensuring the parameters are shared across these inputs. Compared with the former, a unified surrogate model offers the advantage of utilizing all observations' information simultaneously without additional sharing mechanisms. This enhances training efficiency and enables flexible inference of posterior probabilities for all configurations.

\begin{figure*}[tb]
    \centering
    \includegraphics[width=2\columnwidth]{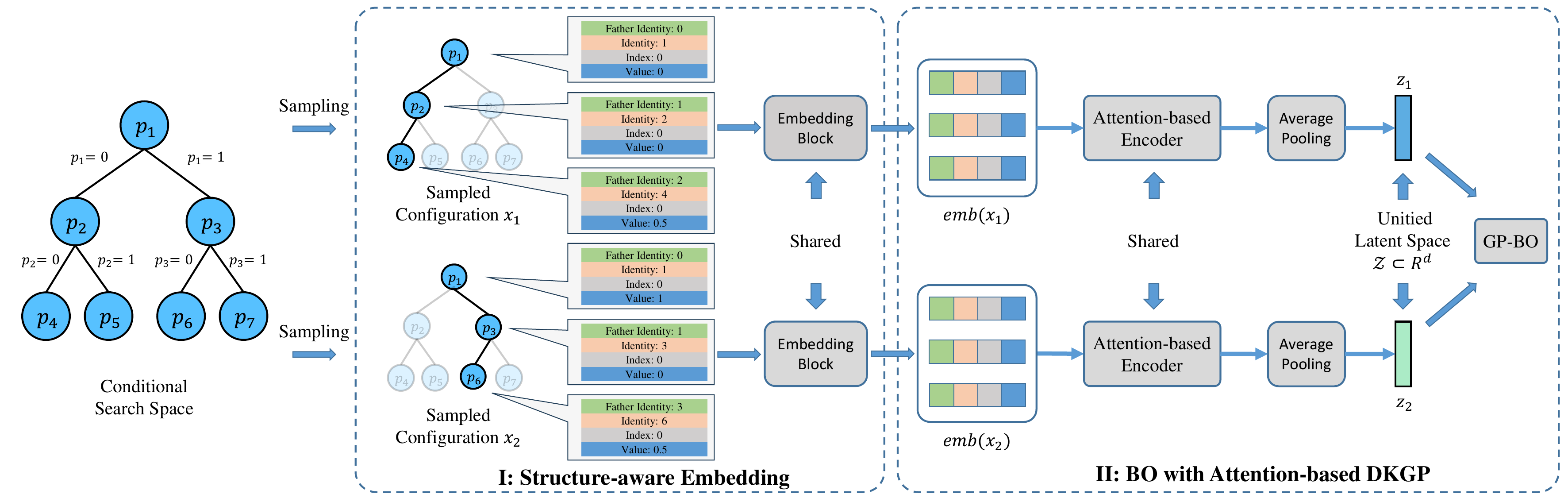}
\caption{The framework of AttnBO. 
We introduce three elements—the identity, index, and father's identity of a hyperparameter, which preserve structural features for each hyperparameter. Each hyperparameter embedding can be considered a token, and a configuration can therefore be viewed as a sequence of tokens. Then we employ an attention-based encoder to capture the relationships of these tokens and project all sequences into a unified latent space where a GP-based BO can work directly.}
\label{fig:framework}
\end{figure*}

\subsection{Structure-aware Embeddings}
\label{embeddings}
The hyperparameters in different subspaces may have different semantics and relationships, which the surrogate model should be aware of when modeling the tree-structured response surface. As Fig.2 shows, there are two sampled configurations $(p_1, p_2, p_4)$ and $(p_1, p_3, p_6)$ from different subspaces. Obviously, the hyperparameters in the two subspaces have different semantics and relationships, making the value of the hyperparameter not enough to represent its feature. 

Therefore, we assign each hyperparameter an embedding that contains semantic and dependency information, instead of only considering the value of the hyperparameter as in previous BO methods~\cite{SMAC,Tree-Structured,addtree_icml}. From the data structure view of this tree structure, since each node has only one parent node or not, we can serialize the tree by storing nodes and their corresponding father nodes. Thus, as Fig.~\ref{fig:framework} shows, we can encode a hyperparameter $p^{x^{i}_{j}}_{k}$ to a structure-aware embedding with four elements:

\textbf{1) The identity embedding $id\_emb(p^{x^{i}_{j}}_{k})$}. It works as an identifier and represents the semantic information of a hyperparameter. First, we adopt ordinal encoding to encode the identities of hyperparameters into numerical codes (1, 2, ...), represented by $id(p^{x^{i}_{j}}_{k})$.  Similar to introducing vectorial meta features~\cite{DBLP:BOHAMIANN_conf/nips/SpringenbergKFH16,DBLP:ablr}, we use a trainable map to get the final identity embedding: $id\_emb(p^{x^{i}_{j}}_{k}) = \phi_{id}(id(p^{x^{i}_{j}}_{k}))$.

\textbf{2) The index embedding $idx\_emb(p^{x^{i}_{j}}_{k})$}. In some NAS problems, some hyperparameters, such as the number of hidden units in a multi-layer deep neural network, may be represented as vectors, the dimension of which depends on the number of layers. In this context, each element of this hyperparameter also needs an identifier. We use the index of each element in the vector $idx(p^{x^{i}_{j}}_{k})$ to identify it and a trainable map to get the final index embedding: $idx\_emb(p^{x^{i}_{j}}_{k}) = \phi_{idx}(idx(p^{x^{i}_{j}}_{k}))$. For the hyperparameter $p^{x^{i}_{j}}_{k}$ containing only a scalar, we set $idx(p^{x^{i}_{j}}_{k}) = 0$.

\textbf{3) The value embedding $value\_emb(p^{x^{i}_{j}}_{k})$}. Besides structural information, the value of a hyperparameter is a crucial feature. We apply a trainable linear map $\phi_{value}$ to construct the value embedding.

\textbf{4) The identity embedding of the father (dependent) hyperparameter $id\_emb(p^{x^{i}_{j}}_{k}\uparrow) = \phi_{id}(id(p^{x^{i}_{j}}_{k}\uparrow))$}. It provides the identifier of a hyperparameter's ancestor, which helps preserve the structural information. For the hyperparameters in the root nodes, we set $id(p^{x^{i}_{j}}_{k}\uparrow) = 0$.

We concatenate these four embeddings as the representation of a hyperparameter $p^{x^{i}_{j}}_{k}$. We refer to all operations involved in obtaining this embedding collectively as the embedding block, denoted by $emb(p^{x^{i}_{j}}_{k})$:
\begin{align} 
& emb(p^{x^{i}_{j}}_{k}) = concat(id\_emb(p^{x^{i}_{j}}_{k}), idx\_emb(p^{x^{i}_{j}}_{k}), \nonumber\\
& value\_emb(p^{x^{i}_{j}}_{k}), id\_emb(p^{x^{i}_{j}}_{k}\uparrow)), k=1, 2, ..., d^{i},
\end{align}
where $d^{i}$ represents the dimension of the flat subspace $\chi^{i}$. Using such embeddings, we transform each hyperparameter into a vector representation that incorporates structural information, thereby enhancing the effectiveness and distinctiveness of the hyperparameter features.
The full embedding of a configuration is represented as: 
\begin{align} 
emb(x^{i}_{j}) = \left[emb(p^{x^{i}_{j}}_{1}), emb(p^{x^{i}_{j}}_{2}),...,emb(p^{x^{i}_{j}}_{d^{i}})\right].
\end{align}
To demonstrate the effectiveness of the structure-aware embedding, we conducted an ablation study on these embeddings, and the experimental results can be found in Fig.~\ref{fig:ablation}.

\subsection{BO with Attention-based DKGP}
\label{AttnBO}

Compared to other surrogate models~\cite{SMAC,TPE}, Gaussian Processes (GPs) offer better uncertainty estimation and higher sample efficiency in practical applications. However, handling configurations that vary in length and contain different hyperparameters is still a challenge for GP. To address this issue, we introduce an attention-based encoder to adapt the GP for this context. This encoder works as a deep feature extractor within the deep kernel framework~\cite{DKL}, allowing us to capture global relationships among hyperparameters and project variable-length configurations into a unified latent space $\mathcal{Z} \subset \mathbb{R}^d$. A GP can then be built on this latent space $\mathcal{Z}$ using a standard kernel function. 

Consider a black-box function with noisy observations $y_{i} = f(x_{i}) + \epsilon, i \subset {1,...,n}, \epsilon \sim \mathcal{N}(0, \sigma)$, we have a dataset $D$ of $N$ noisy observations in a conditional space $\chi$ that has $n$ flat subspaces $\left\{\chi^{1} \cup \chi^{2} \cup ... \cup \chi^{n}\right\}$, $N = \sum_{i=1}^{n} N^{i}$, $D = \left\{D^{1}, D^{2}, ..., D^{n}\right\}$, where $D^{i}$ means all observations $\left\{(x_{j}^{i}, y_{j}^{i})| j = 1,2,...,N^{i}\right\}$ in subspace $\chi^{i}$.
We adopt the deep kernel learning framework~\cite{DKL} to learn the weights of the embedding block, attention-based encoder, and the parameters of the kernel function jointly by maximizing the log marginal likelihood in eq.~\ref{loglikelihood}. In our framework, the deep kernel matrix is as follows:
\begin{align}
\label{eq:Deep Kernel Matrix}
\mathbf{K}_{deep} & = k_{deep}(\mathbf{X}, \mathbf{X} | \theta, \omega) + \sigma^{2} \mathbf{I} \nonumber\\
& = k(\phi(emb(\mathbf{X}, \omega_{1}), \omega_{2}), \nonumber\\
&\quad\quad\;\phi(emb(\mathbf{X}, \omega_{1}), \omega_{2}) | \theta) + \sigma^{2} \mathbf{I},
\end{align}
where $\omega_{1}, \omega_{2}$ are two subsets of $\omega$ representing the weights of the embeddings and the attention-based encoder.

With a well-fitted DKGP, the predictive posterior distribution of the objective function $f$ at $x_{*}$ is as follows:
\begin{align} 
f_{*}& |\mathbf{X}, \mathbf{y}, x_{*} \sim \mathcal{N} \left( \overline{f}_{*}, var(f_{*})) \right),\\
\overline{f}_{*} & = \mathbf{k}_{deep_{*}}^{\top}\mathbf{K}_{deep}^{-1} \mathbf{y}, \\
var(f_{*}) & = k_{deep}(x_{*}, x_{*}) - \mathbf{k}_{deep_{*}}^{\top}\mathbf{K}_{deep}^{-1} \mathbf{k}_{deep_{*}}.
\end{align}
The deep kernel matrix $\mathbf{K}_{deep}$ can be found in eq.~\ref{eq:Deep Kernel Matrix}, and we write $\mathbf{k}_{deep_{*}} = k_{deep}(x_{*}, \mathbf{X})$ to donate the vector of covariances between the test point $x_{*}$ and the training points $\mathbf{X}$. In this paper, we use the Mat\'ern 5/2 kernel function to accommodate the DKGP model and adopt EI acquisition function to choose the next query. During the acquisition stage, we utilize lbfgs optimizer to optimize EI on each subspace and find the most valuable configurations to query in each subspace, enabling parallel Bayesian optimization on the objective functions. For those more complex search spaces that have a large number of subspaces, optimizing the acquisition function within each subspace is computationally expensive. In such cases, we can directly employ random search to optimize the acquisition function over the entire space to give batch queries. Under the sequential BO setting, we choose the configuration that maximizes the acquisition function among all subspaces. The detailed procedure of the algorithm is shown in Algorithm~\ref{alg:attnbo}.

\begin{algorithm}[t]
\newcommand{\nonl}{\renewcommand{\nl}{\let\nl\oldnl}}
\caption{AttnBO: An Attention-based Bayesian Optimization Method for Conditional Search Spaces.} \label{alg:attnbo}
\SetKwInOut{Input}{Inputs}
\Input{A black-box function $f$ defined on a conditional search space $\chi = \chi^{1} \cup \chi^{2} \cup ... \cup \chi^{n}$;\\
The batch size $B (B <= n)$;\\
The number of total training iterations $T$.
}
Randomly sample two initial points $(N^{i} = 2)$ to evaluate from each subspace, resulting in $N=2n$ initial points in total.\\ 
Get the initial dataset: $D_{0} = \left\{(x_{j}^{i}, y_{j}^{i})| i=1,2,...,n, j = 1,2,...,N^{i}\right\}$\\
\For{$t := from~1~to~T$}{
Fit the Deep Kernel Gaussian Process by maximizing the log marginal likelihood (eq.~\ref{loglikelihood}) with Adam optimizer.\\
Optimize the acquisition function in each subspace: $x_{*}^{i} = \mathop{\arg\max}_{x \in \chi^{i}} \alpha(x), i=1,2,...,n.$\\
Get the next queries: $X_{*} = \left\{ x_{b} | b=1,2,...,B \right\} = TopB(\left\{ \alpha(x_{*}^{i}) | i=1,2,...,n \right\})$ \\
Query $f$ and get new observations: $D_{*} = \left\{ (x_{b}, y_{b})|b=1,2,...,B, y_{b} = f(x_{b})\right\}.$\\
Update dataset: $D_{t} = D_{t-1} \cup D_{*}$
}

\textbf{Output:} The best point $x_{opt}$ in history.\\

\nonl $\ddagger$: $TopB$ is a function that returns the top B configurations ranked by the acquisition function.
\end{algorithm}

\section{Experiments}
\subsection{Experimental setting}
\textbf{Tasks.} To demonstrate the efficiency and efficacy of AttnBO, we conduct experiments on multiple tasks:  
For the simulation task, we follow the setting of~\citet{addtree_icml}, and the details of the search space can be found in our supplementary material. We adopt $log_{10}$ distance between the best minimum achieved so far and the practical minimum value as the y-axis. 

Following the solid work~\citet{MNAS} in the NAS field, we set an optimization problem in a complex search space which includes a minimum of 29 and a maximum of 47 hyperparameters depending on different conditions —- the number of the blocks ranging from 4 to 7. There are both categorical and continuous hyperparameters in this NAS space, and the candidate will be evaluated on CIFAR-10 dataset after 100 training epochs. The details of the settings of this search space can be found in the supplementary material.

For the OpenML tasks, we designed two conditional search spaces, one for SVM and one for XGBoost, both of which are widely used machine learning models for tabular data.
Moreover, we also combine the two search spaces via an algorithm variable as Fig.~\ref{fig:new_tree_sp} shows, leading to a CASH problem and making the search space more complex having six subspaces and 15 hyperparameters. The details of these search spaces are shown in our supplementary material. Supported by OpenML~\cite{OpenML2013}, we consider 6 most evaluated datasets: [10101, 37, 9967, 9946, 10093, 3494].

Benefiting from the capability to handle configurations across various subspaces, our surrogate model enables large-scale meta-learning on multiple source tasks from various search spaces. We verify this feature on HPO-B-v3~\cite{HPOB}, a large-scale hyperparameter optimization benchmark that contains a collection of 935 black-box tasks for 16 hyperparameter search spaces evaluated on 101 datasets. We set the ID of a search space as the father node of its hyperparameters, resulting in a tree-structured search space. We meta-train our model on all training data points of the 16 search spaces and fine-tune it on the test tasks to get the final performance.

\paragraph{Baselines.}
We compare AttnBO with Random Search~\cite{DBLP:rs_journals/jmlr/BergstraB12} and four BO baselines for the conditional space on all tasks, including two GP-based methods (Bandits-BO~\cite{bandits-bo}, AddTree~\cite{addtree_icml, addTree_tpami}) and two non-GP methods (SMAC~\cite{SMAC}, TPE~\cite{TPE}). Compared to the naive independent GPs, Bandits-BO is more advanced and can easily degrade to the latter. Therefore, we choose it as our baseline rather than the naive independent GPs. Moreover, we also compare with Bandits-BO under a parallel setting on real-world OpenML tasks to demonstrate our ability of batch optimization. The implementation details of these baselines can be found in our supplementary material.

\paragraph{Implementation details.}
We adopt the Transformer encoder as the deep kernel network to project the configurations in different subspaces into a unified latent space $\mathcal{Z}$. Specifically, we employ 6 attention blocks with 2 parallel attention heads. The dimensionality of input and output is dmodel = 256 (4 $\times$ 64), and the inner layer also has a dimensionality of 512. 
We adopt average pooling to integrate the output of the transformer encoder and utilize a multi-layer perceptron (MLP) with 4 hidden layers, which has [128, 128, 128, 32] units of each hidden layer, to project the features of the configurations into 32-dim vectors. The parameters of the encoder are selected based on their performance on the SVM and XGBoost tasks.
We train the embedding layer and attention-based encoder jointly for 100 epochs using Adam~\cite{adam}. We set the initial learning rate to 0.001 and reduce it by half every 30 epochs. More details of our implementation can be found in our supplementary material.

\paragraph{Other settings.}
Following the settings of Bandits-BO, we give $2n$ random points to initialize BO methods. Then, we run BO on the simulation and OpenML tasks until 80 observations (without initial points) are collected and repeat the experiment 10 times to reduce the impact of random seeds. 
For the NAS tasks, we train each candidate on CIFAR-10 training set for 100 epochs and evaluate on the testing set. Because the evaluation of a configuration in this task is very expensive, we only repeat the experiment 3 times.

\subsection{Experiment Analysis}
\paragraph{Simulation Function.}

\begin{figure}[H]
\centering
\includegraphics[width=0.65\linewidth]{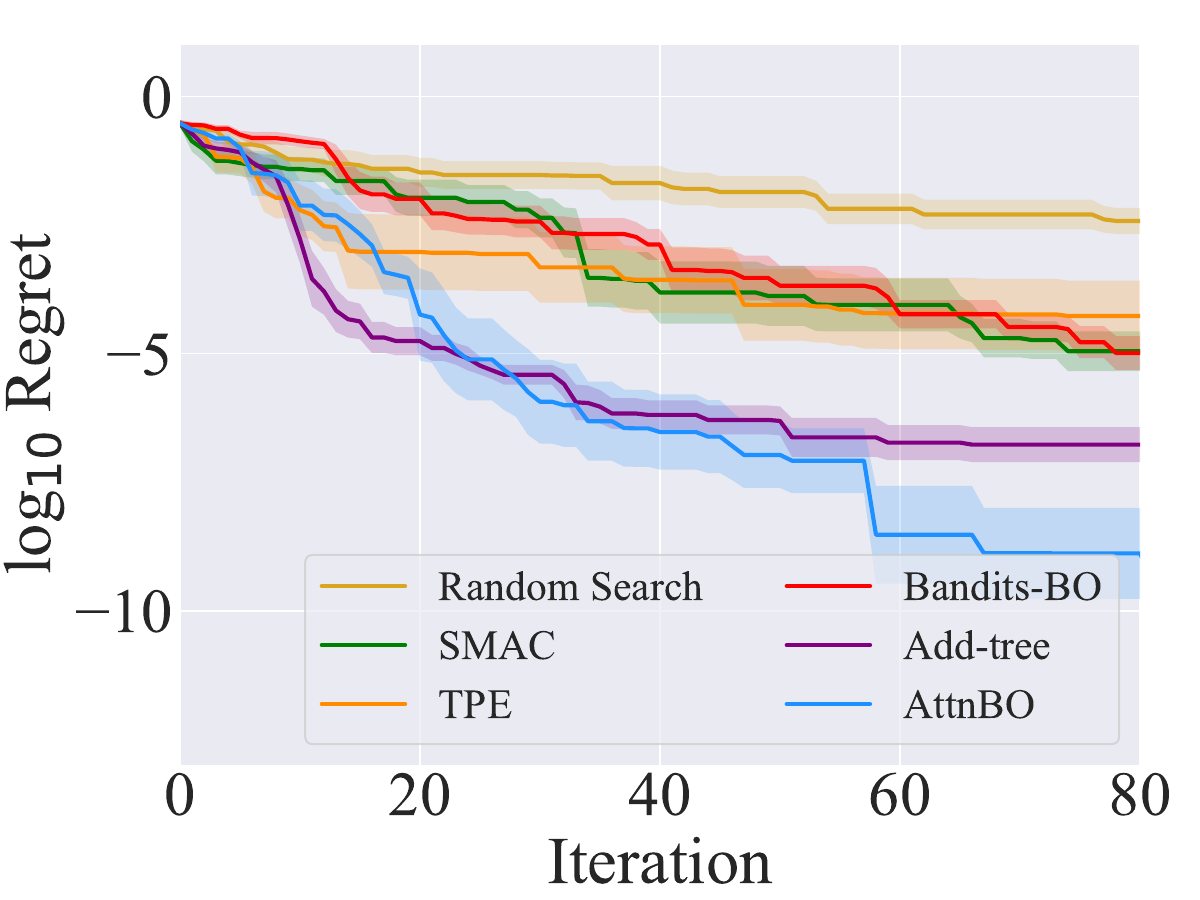}
\caption{Performance of our AttnBO and baselines on the conditional simulation objective function.}
\label{fig:simfunc}
\end{figure}

Following the setting of~\citet{addtree_icml}, we compare our AttnBO with other baselines on this additive structure objective function. As shown in Fig.~\ref{fig:simfunc}, our method performs best on this simulation function. Here, for a fair comparison, we reimplement the experiment and set the same random seeds as all other algorithms. (Probably, we did not get the same results as shown in their paper due to the different number of initial points.) 

\begin{figure}[H]
  \includegraphics[width=0.9\columnwidth]{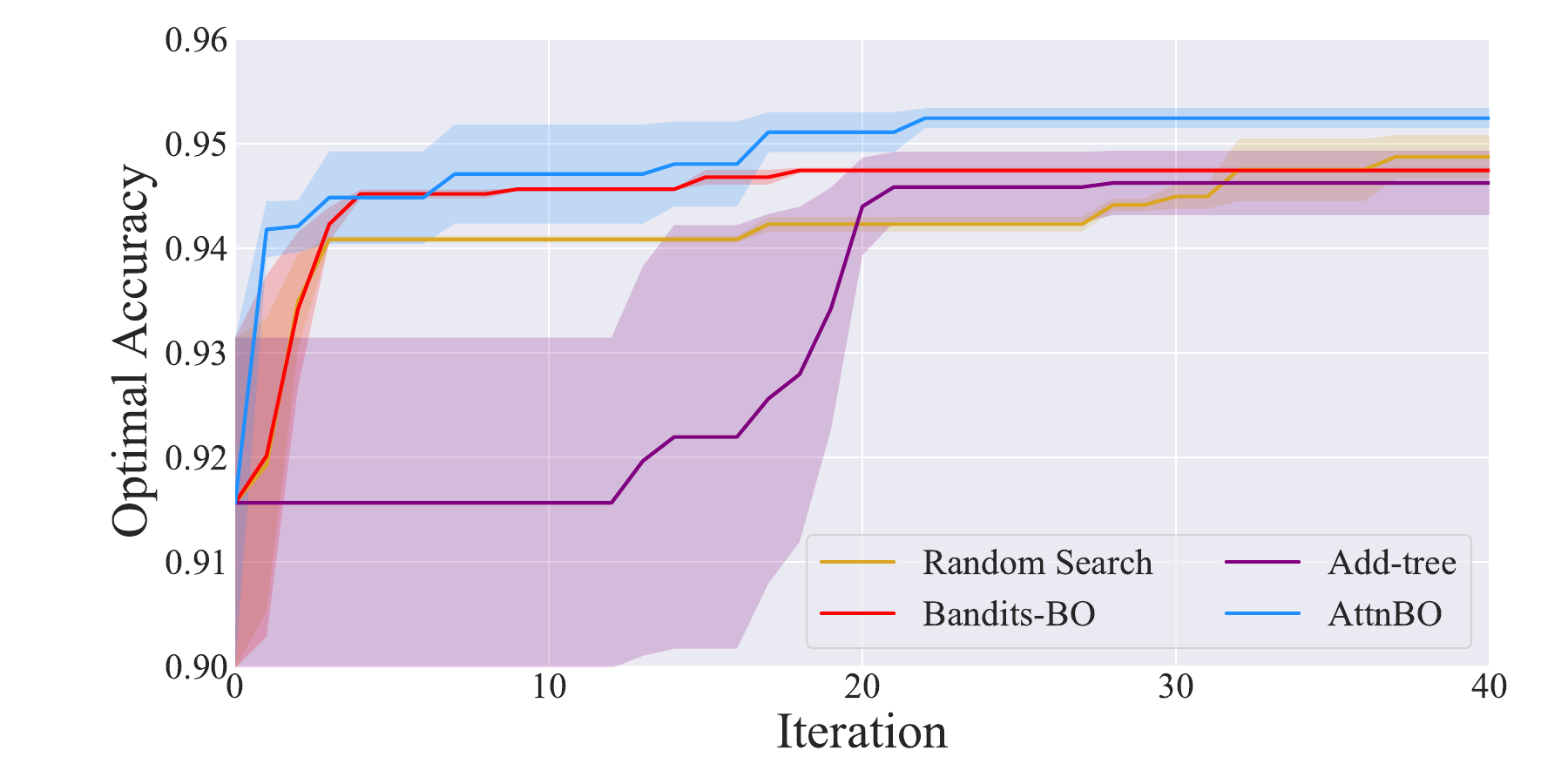}  

\caption{Performance of baselines and AttnBO on the complex NAS space.}
\label{fig:NAS}
\end{figure}

\paragraph{Neural Architecture Search.}
Considering the evaluation of a deep neural network is expensive, parallelization becomes especially important and necessary, which could improve the efficiency of the optimization process. However, the state-of-the-art method AddTree~\cite{addtree_icml} doesn't support parallelization, which will still give one query per BO iteration in this experiment. We show the optimal accuracy after each BO iteration for all methods in Fig.~\ref{fig:NAS}. 

With more hyperparameters and more complex condition settings, the ability to explore becomes crucial. Compared to other methods, Add-Tree is limited in its capability to explore or exploit various configurations within a BO loop due to the inability to provide batch queries. As a result, the opportunity for observation is reduced, leading to a failure in finding optimal configurations during the early stages. On the other hand, Random Search demonstrates better performance on this task because of its strong ability to explore across each dimension in a larger space with parallelism~\cite{DBLP:rs_journals/jmlr/BergstraB12}.
In contrast to existing methods, our AttnBO has the advantage of exploiting the relationships between hyperparameters, which allows us to learn better representations of configurations.
Additionally, AttnBO also enables parallel optimization by selecting the best candidate in each subspace, leading to higher efficiency throughout the BO process.

\begin{figure*}[tb]
    \centering
    \includegraphics[width=0.75\linewidth]{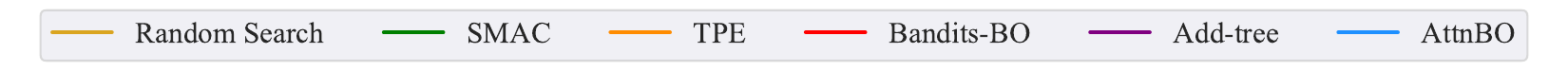}
    
  \subfigure[SVM]{
    \label{fig:svm}
    \includegraphics[width=0.5\columnwidth]{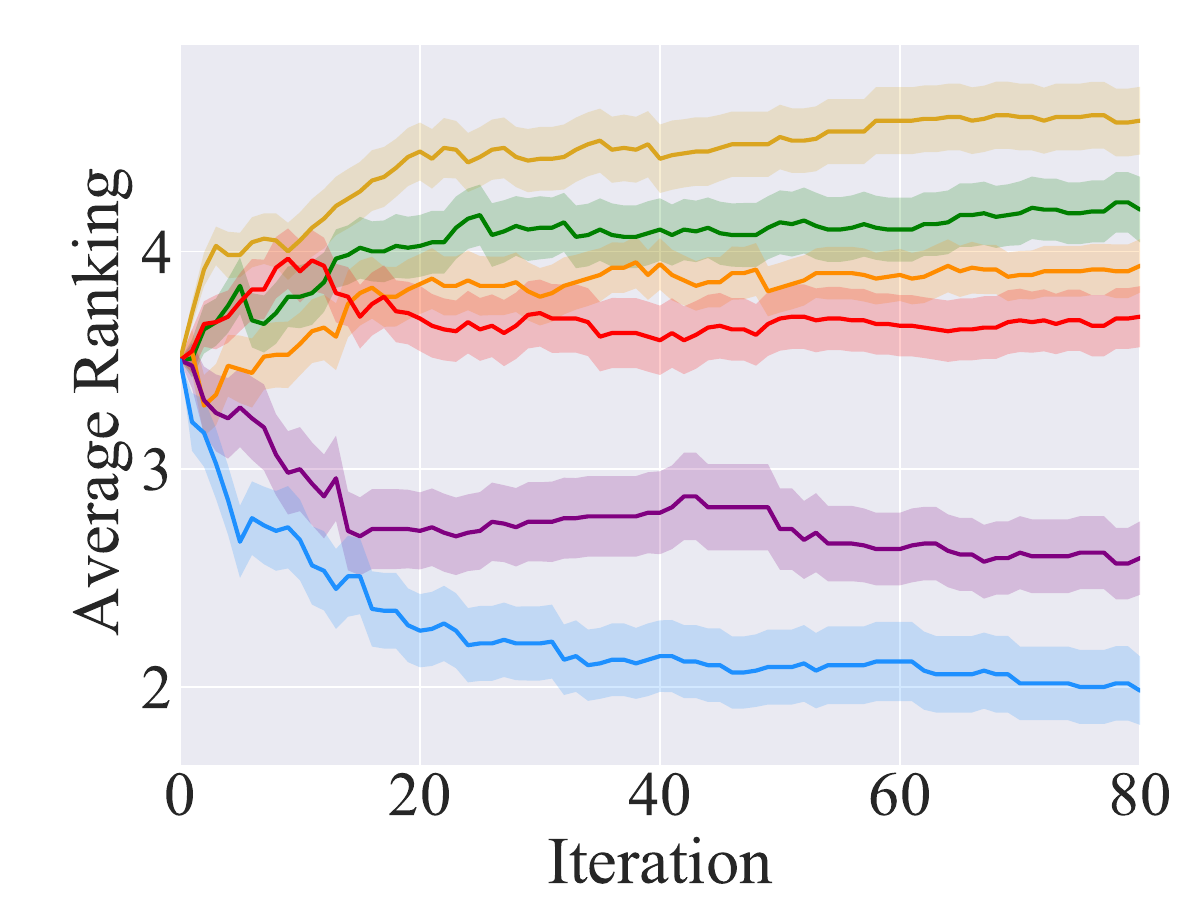}  
    }
  \subfigure[XGBoost]{
    \label{fig:xgb}
    \includegraphics[width=0.5\columnwidth]{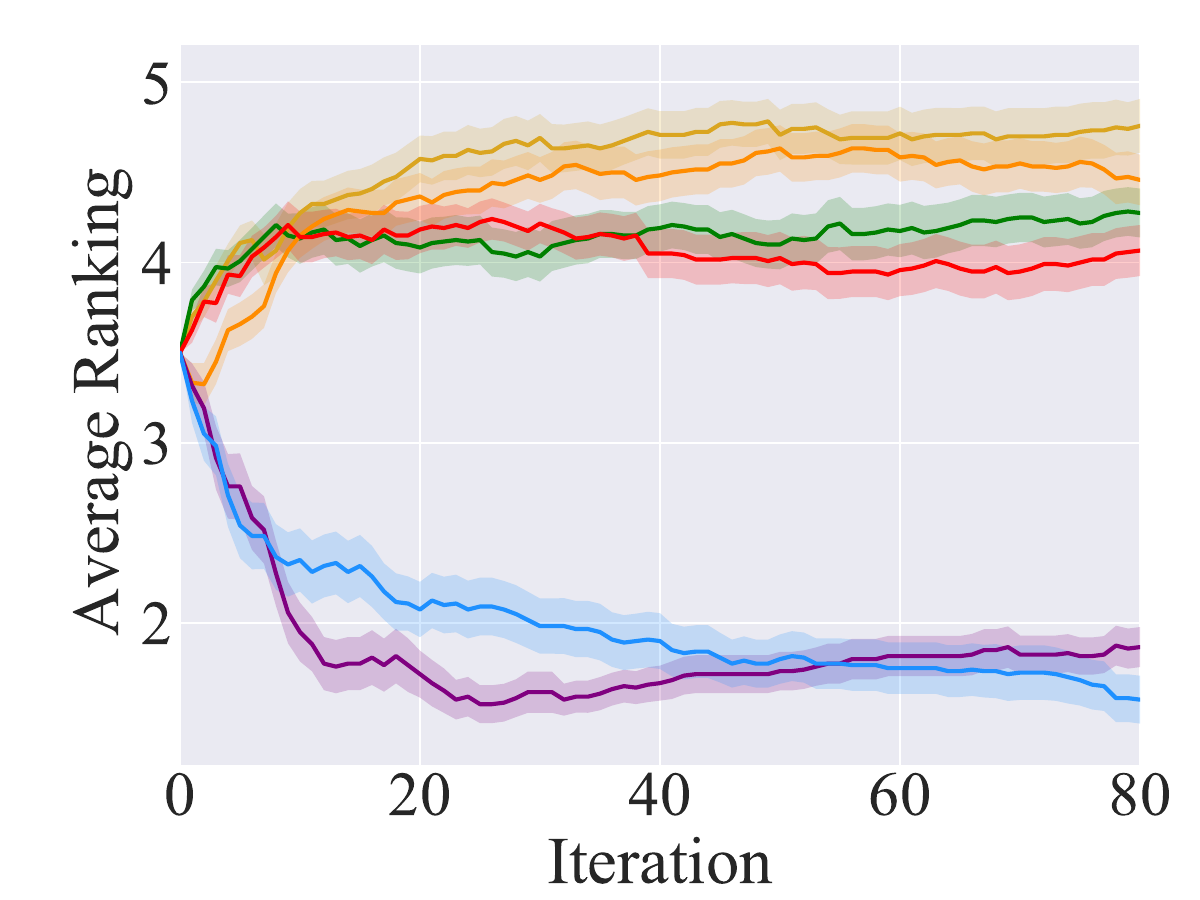}  
    }
  \subfigure[SVM + XGBoost]{
    \label{fig:svm_xgb}
    \includegraphics[width=0.5\columnwidth]{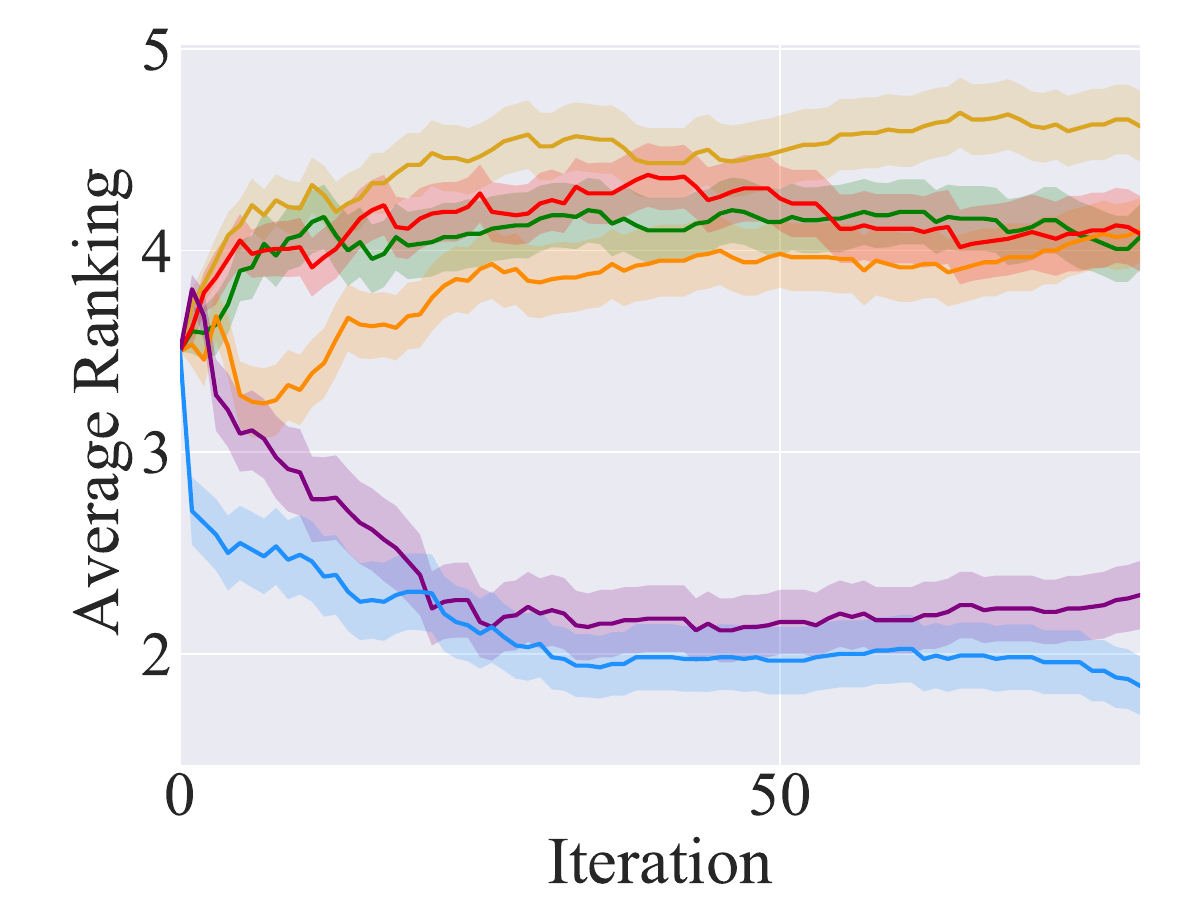}  
    }
    
\caption{Average rankings of various methods on three machine-learning tasks evaluated on real-world OpenML datasets.}
\label{fig:openml}
\end{figure*}

\paragraph{Real-world tasks on OpenML.}
Fig.~\ref{fig:openml} reports the average ranking of performance on three hierarchical search spaces of two machine-learning models, which were evaluated on 6 real-world datasets randomly selected from OpenML~\cite{OpenML2013, OpenMLPython2019}. Our proposed method achieves the best performance on all three tasks and, in particular, outperforms the start-of-the-art BO method AddTree~\cite{addtree_icml} for conditional search spaces. We also report the performance of all baselines on each dataset, which can be found in our supplementary material.

\begin{figure}[H]
    \centering
    \includegraphics[width=1.0\linewidth]{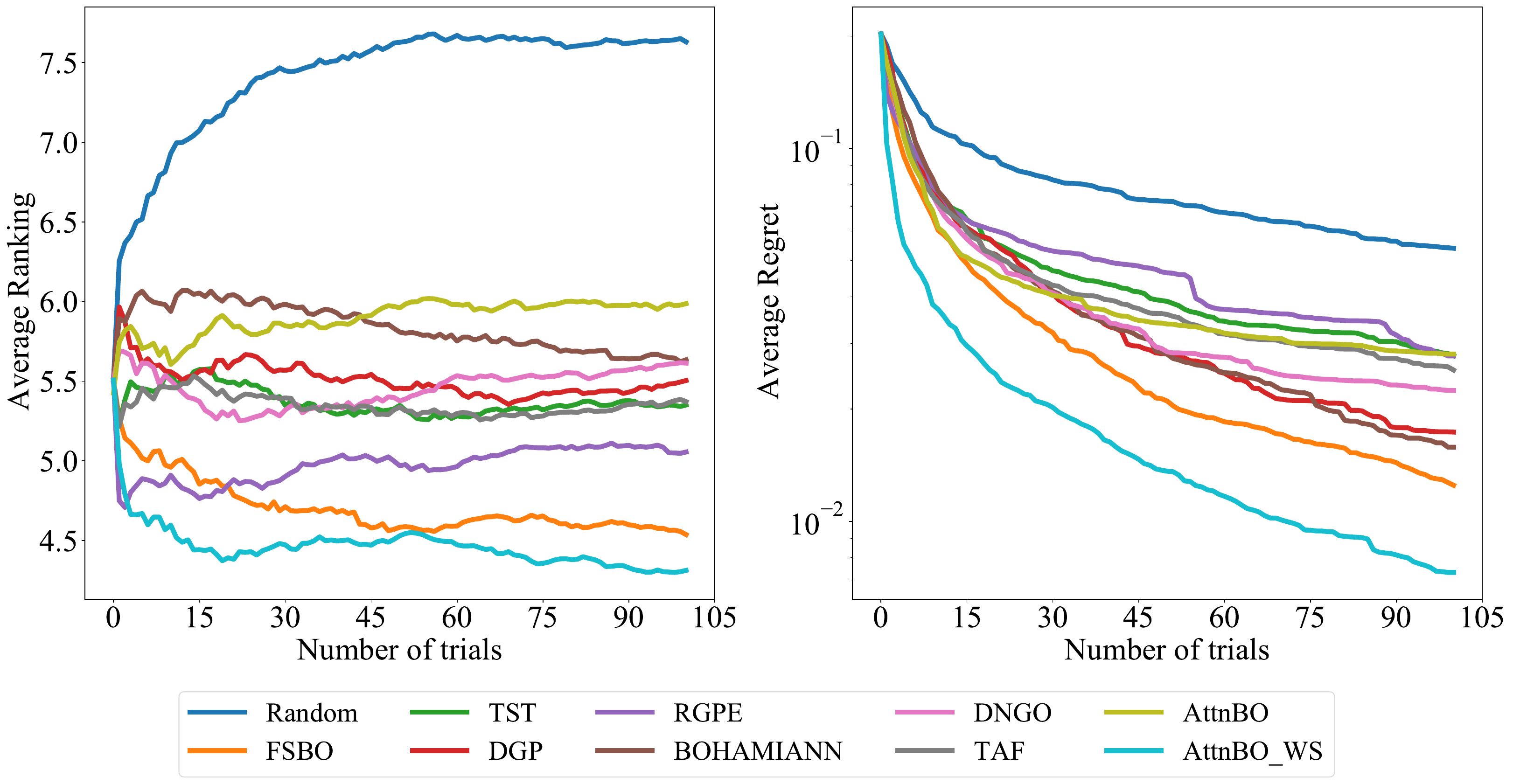}

\caption{Performance of various black-box optimization methods on HPO-B benchmark.}
\label{fig:hpob}
\end{figure}

\paragraph{HPO-B Benchmark.}
Fig.~\ref{fig:hpob} displays the performance of our method compared to other default baselines as reported in~\cite{HPOB}. We also compare AttnBO with and without warm-starting, denoted as AttnBO\_WS and AttnBO, respectively. The results demonstrate the effectiveness of the warm-starting phase on large-scale datasets. Besides, with warm-starting on HPO-B, our proposed AttnBO can achieve competitive performance with state-of-the-art transfer learning methods.

\paragraph{Computational Cost}
Compared to other baselines, our method incurs slightly higher computational costs due to the training of the deep encoder. Table~\ref{tab:time_cost} summarizes the total time (in minutes) required by our method and two other BO methods to complete 100 iterations on the simulation task. Since the simulation function involves no evaluation cost, the recorded time is effectively equivalent to the algorithm’s runtime. The results indicate that our method takes only about 5 minutes longer than the other two methods. For objective functions with significantly high evaluation costs, this additional overhead becomes negligible.

\begin{table}[H]
\centering
\resizebox{0.4\textwidth}{!}{
\smallskip
\begin{tabular}{lcccccc}
\toprule

\multicolumn{3}{c}{\textbf{Time Cost (minutes)}}\\
 \textbf{$AttnBO$} & \textbf{$Bandits-BO$} & \textbf{$Add-tree$} \\
\midrule

17.05 & 
11.96 & 
12.73 \\

\bottomrule
\end{tabular}
}

\caption{The total time cost over 100 iterations on the simulation task.}
\label{tab:time_cost}
\end{table}

\paragraph{Sample Efficiency.} 
Compared to separate GP models, the unified model exhibits higher sample efficiency. To illustrate the higher sample efficiency of the unified model compared to separate GPs, we conducted a comparison based on the simulation function. The results show that our method can achieve the same performance within 100 observation points that the separate models require 200 points to reach. Thus, our method will become more advantageous when the observation cost of the black box is exceedingly expensive.

\begin{figure}[H]
\centering
\includegraphics[width=0.65\linewidth]{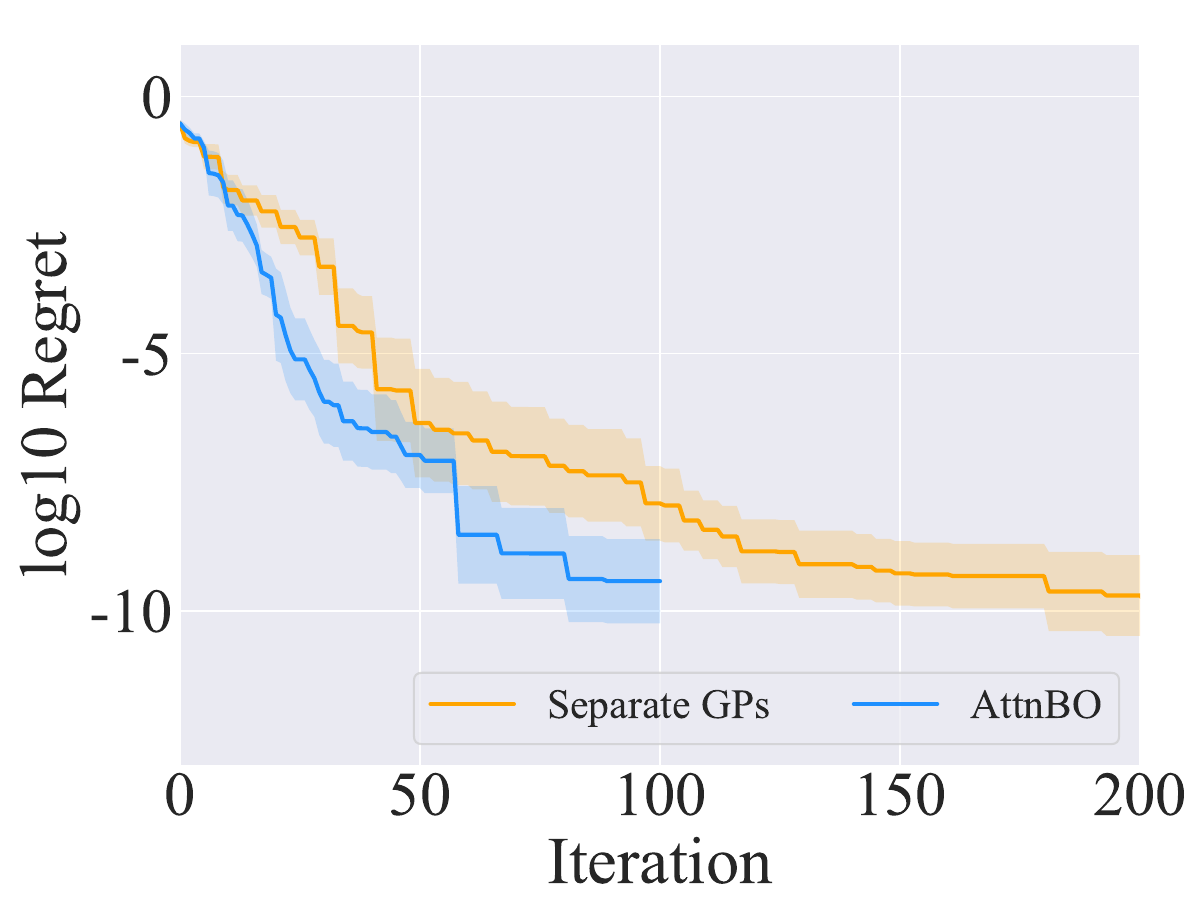}
\caption{Comparison of AttnBO with the Separate GP Approach on the conditional simulation objective function.}
\label{fig:sgp_simfunc}
\end{figure}

\subsection{Ablation Study}
\label{AblationStudy}

\begin{figure}
    \centering
    \includegraphics[width=1\columnwidth]{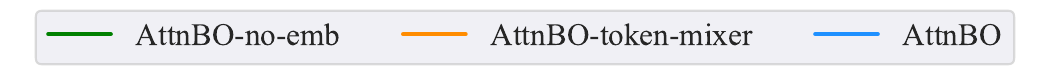}

  \subfigure[SVM]{
    \label{fig:svm_ab}
    \includegraphics[width=0.46\columnwidth]{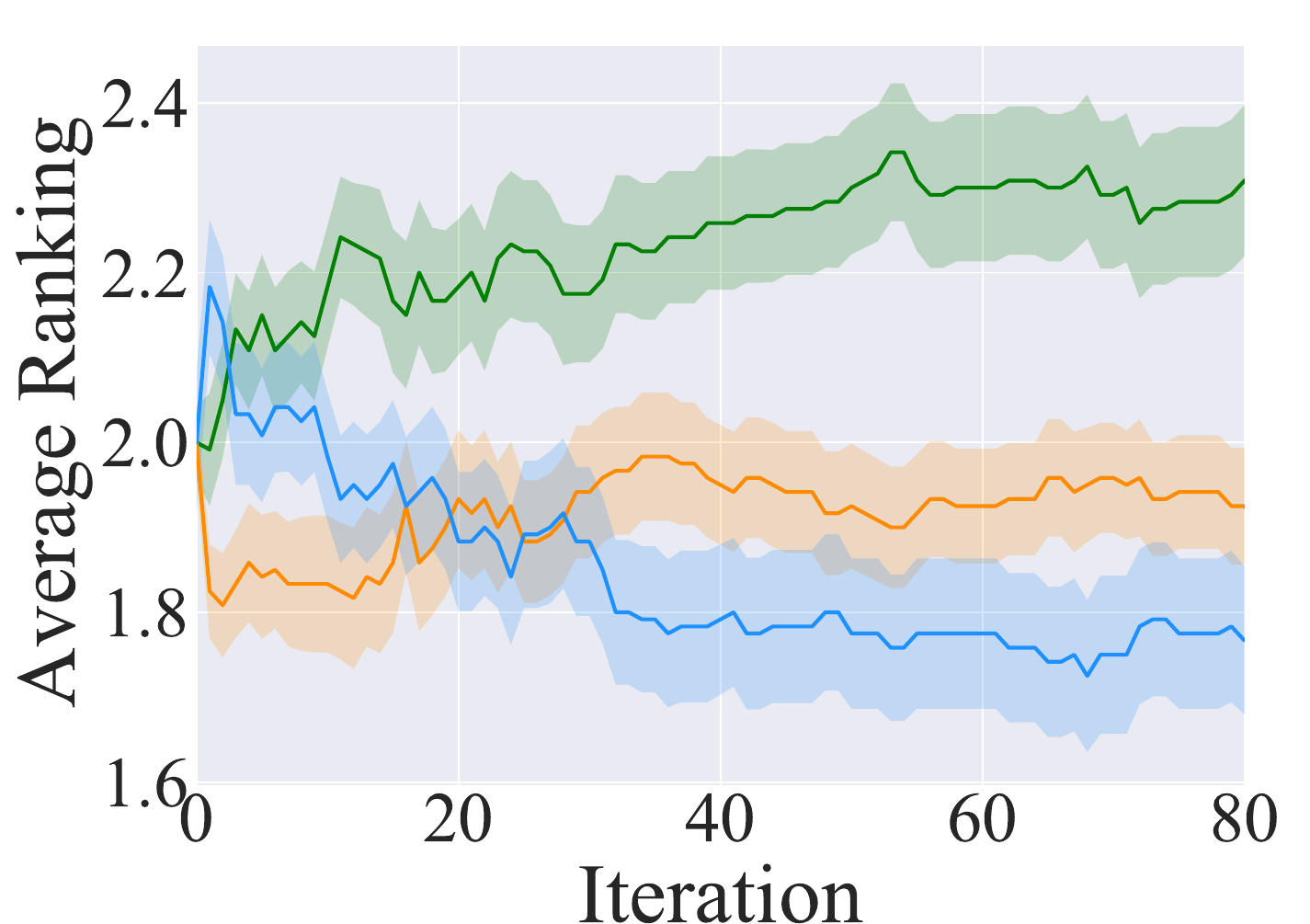}  
    }
  \subfigure[XGBoost]{
    \label{fig:xgb_ab}
    \includegraphics[width=0.46\columnwidth]{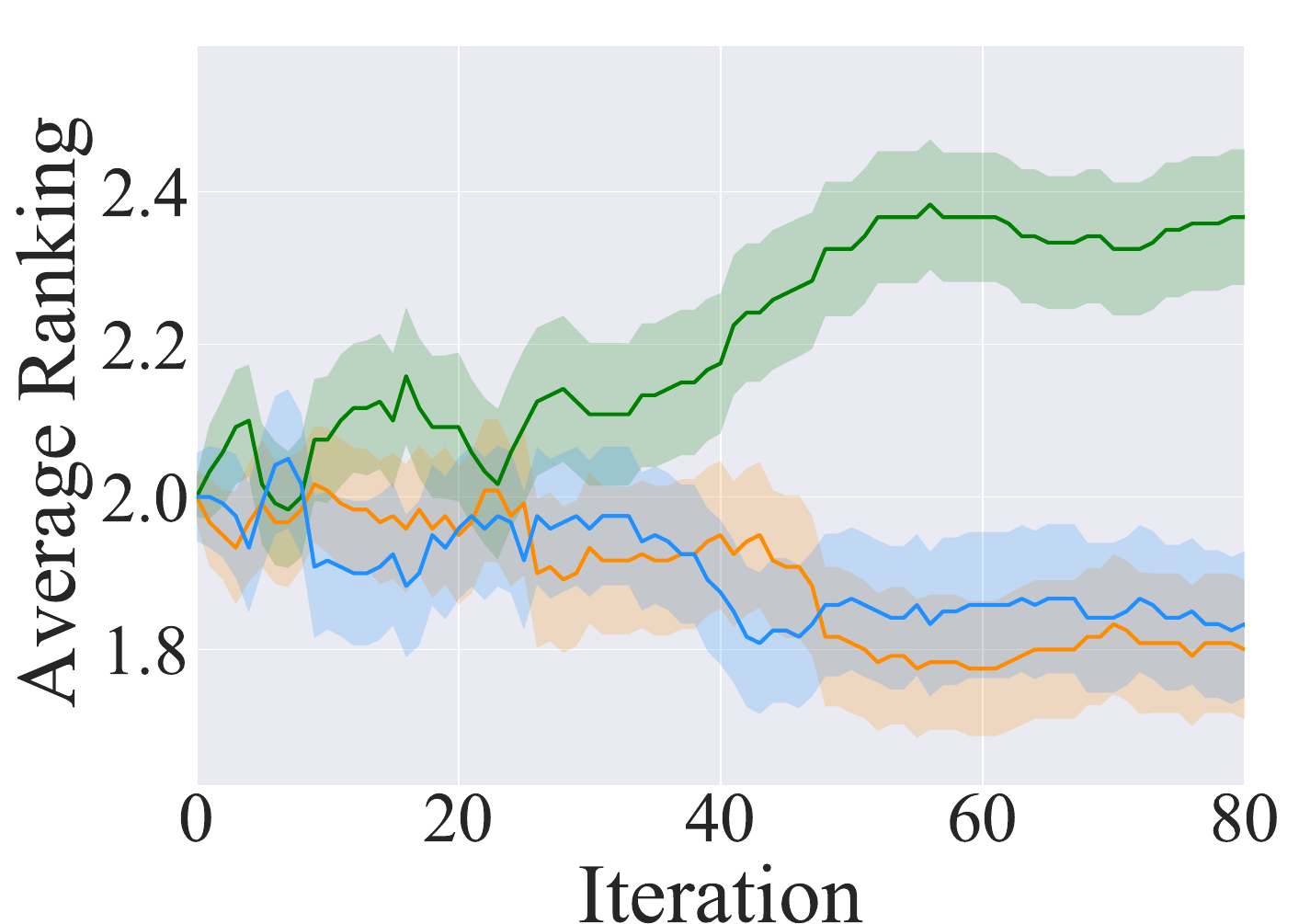}  
    }

\caption{Performance of our AttnBO and two variants on the SVM and XGBoost task.}
\label{fig:ablation}
\end{figure}

\paragraph{Structure-aware Embedding.} In this experiment, we compare our method with two variants, AttnBO-no-emb and AttnBO-token-mixer, on two machine-learning tasks. On the one hand, to verify the effectiveness of the embedding we proposed, we drop the structure-aware embedding and only use the values of the hyperparameters as their features. We denote this method as AttnBO-no-emb. On the other hand, instead of using the average pooling operator for feature fusion, we introduce an additional token to represent the global feature of the sequence, which is often used in the Computer Vision field~\cite{VIT} and denoted as AttnBO-token-mixer. Fig.~\ref{fig:ablation} shows the average ranking of these methods over 5 datasets in each search space.
Obviously, our proposed embedding method helps to capture the relationships between hyperparameters and leads to higher effectiveness.

\paragraph{Architecture of the Attention-based Encoder.} 
This experiment is to explore the architecture of the attention-based encoder. 
The number of the self-attention blocks and parallel heads are denoted as $n_a$ and $n_b$, respectively. 
Table~\ref{tab:architect} shows all candidates' average regret and ranking on these two tasks. The results show that changes in the network structure had minimal impact on average accuracy performance, with the final average rankings of different structures being very similar. This indicates that our method is not sensitive to slight variations in network architecture.
We selected the configuration with the lowest combined average ranking across the two tasks ($n_a=6,n_b=2$, with MLP) as our default network structure.

\begin{table}[t]
\centering
\resizebox{0.475\textwidth}{!}{
\smallskip
\begin{tabular}{lcccccc}
\toprule

\multicolumn{3}{c}{\textbf{Architecture}}& \multicolumn{2}{c}{\textbf{SVM}}&  \multicolumn{2}{c}{\textbf{XGBoost}}\\
 \textbf{$n_a$} & \textbf{$n_b$} & \textbf{MLP} & \textbf{accuracy} & \textbf{ranking} &  \textbf{accuracy} & \textbf{ranking}\\
\midrule

3 & 
8 & 
$\times$ &
0.84$\pm$0.05 &
4.73$\pm$0.24 &
0.88$\pm$0.04 &
4.71$\pm$0.28 \\

3 & 
8 & 
\checkmark  &
0.85$\pm$0.05 &
4.81$\pm$0.25 &
0.89$\pm$0.04 &
5.12$\pm$0.27 \\

4 & 
6 & 
$\times$  &
0.84$\pm$0.05 &
4.80$\pm$0.25 &
0.89$\pm$0.04 &
4.21$\pm$0.28 \\

4 & 
6 & 
\checkmark  &
0.85$\pm$0.05 &
4.41$\pm$0.26 &
0.89$\pm$0.04 &
4.68$\pm$0.24 \\

5 & 
4 & 
$\times$  &
0.85$\pm$0.05 &
4.45$\pm$0.24 &
0.90$\pm$0.04 &
3.99$\pm$0.28 \\

5 & 
4 & 
\checkmark  &
0.85$\pm$0.05 &
4.07$\pm$0.25 &
0.89$\pm$0.04 &
4.48$\pm$0.29 \\

6 & 
2 & 
$\times$  &
0.85$\pm$0.05 &
4.53$\pm$0.25 &
0.88$\pm$0.04 &
4.72$\pm$0.28 \\

6 & 
2 & 
\checkmark  &
0.84$\pm$0.05 &
4.20$\pm$0.22 &
0.90$\pm$0.04 &
4.10$\pm$0.28 \\

\bottomrule
\end{tabular}
}

\caption{The average accuracy and ranking of candidate network architectures on the SVM and XGBoost tasks. The signs "$\times$" and "$\checkmark$" represent the architecture with and without a MLP after the attention block, respectively.}
\label{tab:architect}
\end{table}

\section{Conclusion}
In this paper, we explore how to model the response surfaces within conditional search spaces in one for efficient optimization in these spaces. We proposed a novel attention-based BO framework AttnBO. Concretely, we proposed a hyperparameter embedding method that can introduce the semantic and dependency information into the feature of each hyperparameter. Then we utilize an attention-based encoder to model the relationships among hyperparameters and project the configurations from different subspaces into a unified latent space. With the powerful attention-based encoder, we build a single standard GP model in the latent space and train the parameters of the deep kernel by the negative log marginal likelihood. Moreover, our proposed method can give a batch of queries in a BO iteration, which improves efficiency when dealing with expensive objective functions. We conduct the experiments on multiple tasks and give sufficient experimental results to demonstrate the effectiveness of our method.

\section{Acknowledgments}
This work was partially supported by the Major
Science and Technology Innovation 2030 "New
Generation Artificial Intelligence" key project
(No. 2021ZD0111700). The authors are grateful to the anonymous reviewers for their insightful
comments and careful proofreading.

\newpage

\bibliography{aaai25}

\newpage

\clearpage

\appendix

\section{\LARGE Supplementary Material}
\section{Implementation Details}
In this section, we'll go into the technical details of our AttnBO and other baselines.
\label{implement}

\subsubsection{AttnBO}
\label{attn_implementation}
\subsubsection{Structure-aware Embeddings}
We use sequential coding to encode the identity of each hyperparameter in a full hierarchical space $\chi$. For example, assume that we have a search space that has three hyperparameters $p_1$, $p_2$, and $p_3$, and $p_2$ is a child of $p_1$. We create a map to encode $p_1$'s identity into 1, $p_2$'s identity into 2, and $p_3$'s identity into 3. Then, we find the identity code of each hyperparameter's father to introduce the dependencies information. Combine the father's identity code and its own identity code, we can get such codes for the three hyperparameters: $p_1: [0, 1]$, $p_2: [1, 2]$, $p_3: [0, 3]$, where code 0 is the padding code for representing a hyperparameter without any dependencies. 

Considering some hyperparameters are vectors and have several dimensions, we introduce the index information of the hyperparameter. For example, we have two hyperparameters in a neural network search space, which are the number of layers $nums\_layer$ and the number of units per layer $nums\_unit$ respectively. Specifically, $nums\_layer$ is the father node of $nums\_unit$ and ranges from 4 to 7, which indicates that the hyperparameter $nums\_units$ will be a vector whose length ranges from 4 to 7 depending on the value of $nums\_layer$. In such a situation, we need to identify each dimension because each dimension in this vector has the same name and father node. For example, if $nums\_layer$ gets 4, then we can get code 1, 2, 3, 4 for each dimension of $nums\_units$. In addition, if a hyperparameter is a scalar and only has one dimension, we use code 0 to represent its index.

Based on these codes, we utilize an embedding layer to get the $id\_emb$ and the father's $id\_emb$, and another embedding layer to get $idx\_emb$ for each hyperparameter. Specifically, we utilize 'nn.Embedding' provided in PyTorch to get the embeddings, which have 64 dimensions in our setting. When we sample a configuration in the search space, we use a linear layer to transform the value of each hyperparameter into a 64-dim vector and concatenate these three embeddings as the representation of each hyperparameter. Then, we can get the full embedding of the configuration as eq.3 and eq.4 show.

\subsubsection{Attention-based Encoder}
We adopt the Transformer encoder as the deep kernel network to project the configurations in different subspaces into a unified latent space $\mathcal{Z}$. Specifically, we employ 6 attention blocks with 2 parallel attention heads. The dimensionality of input and output is dmodel = 256 (4 $\times$ 64), and the inner layer also has a dimensionality of 512. 
We adopt average pooling to integrate the output of the transformer encoder and utilize a multi-layer perceptron (MLP) with 4 hidden layers, which has [128, 128, 128, 32] units of each hidden layer, to project the features of the configurations into 32-dim vectors. In our ablation study, following~\citet{VIT}, we utilize another way to integrate the features of the transformer encoder via an extra token, which we named AttnBO-token-mixer in this paper.

\subsubsection{Deep Kernel Gaussian Process}
For the Gaussian Process model, we utilize Mat\'ern 5/2 as the kernel function and set the mean prior to zero. We adopt the Adam optimizer to train the parameters of the kernel by maximizing the log-likelihood, embedding layer, and attention-based encoder for 100 epochs. We set the learning rate to 0.001 with a decay rate of 0.5 every 30 epochs. For the acquisition, we utilize EI to balance the exploration and exploitation and utilize the lbfgs optimizer to optimize EI in each subspace during the acquisition stage. Unfortunately, a large number of subspaces will make it impossible to optimize EI in each subspace using lbfgs, which performs best in our experiment. In this situation, we can use Thompson sampling as the acquisition to find the next query like~\citet{bandits-bo}. If you still want to use EI, you can just simply use random sampling to optimize EI in the full search space.

\subsection{Baselines}
\label{baselines}
In this section, we provide the specific details of each baseline mentioned in the paper:
\paragraph{Random Search (RS).} Following the description in~\cite{DBLP:rs_journals/jmlr/BergstraB12}, we sample candidates uniformly at random.

\paragraph{Tree Parzen Estimator (TPE).}~\citet{TPE} adopt kernel density estimators to model the probability of configurations with bad and good performance respectively. We use the default settings provided in the hyperopt package (\url{https://github.com/hyperopt/hyperopt}).

\paragraph{SMAC.}~\citet{SMAC} adopt random forest to model the response surface of the black-box function. When dealing with the search space with dependencies, SMAC imputes the inactive hyperparameters in each subspace with default values. We use the default settings given by the scikit-optimize package (\url{https://github.com/scikit-optimize/scikit-optimize}) and impute the default values as SMAC3 package (\url{https://github.com/automl/SMAC3}).

\paragraph{Bandits-BO.}~\citet{bandits-bo} builds a sub-GP in each subspace and uses a Thompson sampling scheme that helps connect both multi-arm bandits and GP-BO in a unified framework. We implement this method in our own framework. For each sub-GP, we use the same settings as our AttnBO except for the deep neural network. We use the Mat\'ern 5/2 as the kernel function and fit the sub-GPs using slice sampling.

\paragraph{AddTree.}~\citet{addtree_icml} proposed an Add-Tree covariance function to capture the global response surface using a single GP, which is the state-of-the-art BO method for the hierarchical search spaces. We use the default settings provided by \url{https://github.com/maxc01/addtree}.

\section{Details of the Benchmarks and Experiments}
To better display the search space with dependencies, we define a YAML format to represent the search space. Following~\citet{omniforce}, we adopt the keywords "type" and "range" to represent the type and domain of the hyperparameter respectively. In addition, we also define the keyword "submodule" to indicate the dependencies among hyperparameters. As for dependencies, in this search space format, we support two types. When the number or distribution of one parameter depends on another parameter, we can use the keyword "submodule" to indicate the relationship between these parameters. For the type of each hyperparameter, we support choice, int, and float for the categorical, integer, and decimal hyperparameters respectively. As to the range of integer hyperparameters, we adopt the left-closed and right-open intervals to represent. For example, if an integer hyperparameter $x_1$ has the range [0...2], it can be 0 or 1. 
For every search space, we will give both the YAML-style and figure-style representation. 

\subsection{Simulation Benchmark}
\begin{figure}[h]
    \centering   
    \includegraphics[width=1\columnwidth]{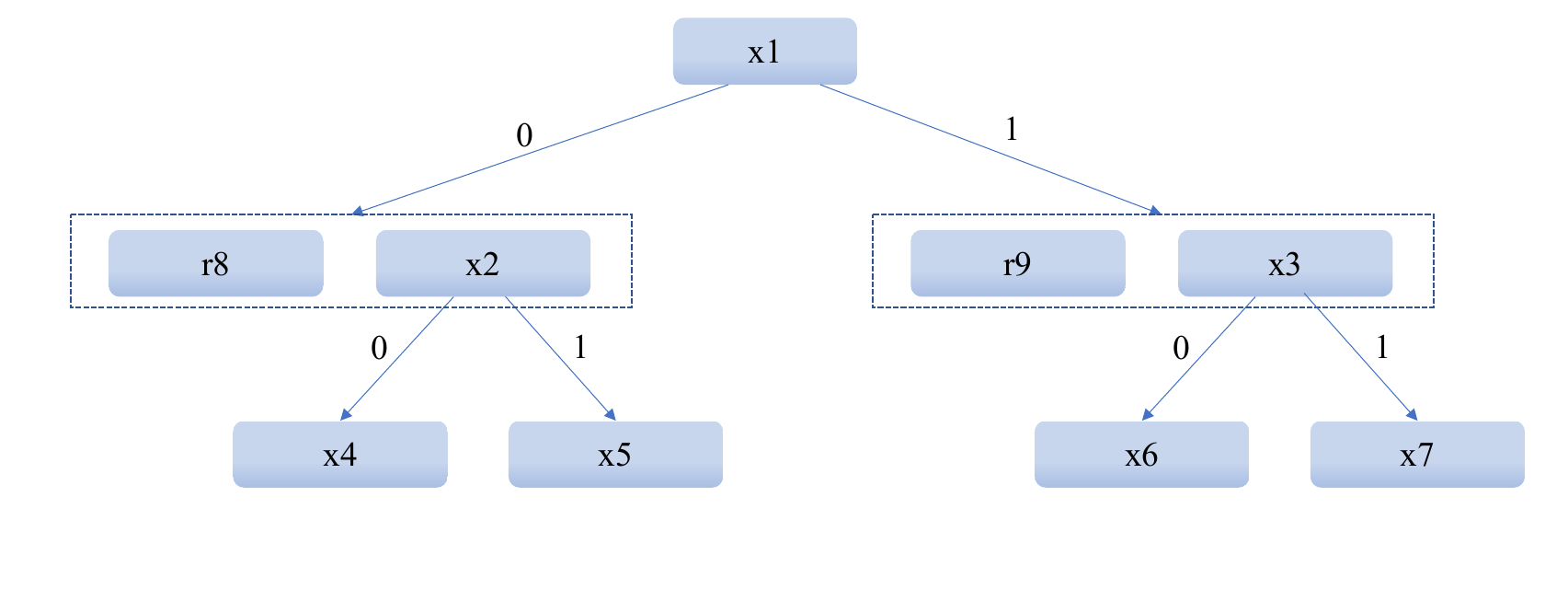}  
\caption{The tree-structured search space on the simulation function presented in~\cite{Tree-Structured}.}
\label{fig:tree_jn}  
\end{figure}

The tree-structure search space of the simulation function that was originally presented in \citet{Tree-Structured} consists of 9 hyperparameters as Listing~\ref{list:simulation_space} and Fig.~\ref{fig:tree_jn} shows. This space has three binary decision variables $x_{1}, x_{2}, x_{3}$, two shared variables $r_{8}, r_{9}$ bound in [0, 1], and four non-shared numerical variables $x_{4}, x_{5}, x_{6}, x_{7}$ bounded in [-1, 1].

\begin{lstlisting}[title={\bf Listing: YAML-style representation of the simulation search space}, label={list:simulation_space}, float=tb, caption={YAML of the simulation search space.}, language=yaml]

x1:
  type:  choice
  range: {0, 1}
  submodule:
    0:
      r8:
        type:  int
        range: [0...2]
      x2:
        type:  choice
        range: {0, 1}
        submodule:
          0:
            x4:
              type:  float
              range: [-1...1]
          1:
            x5:
              type:  float
              range: [-1...1]
    1:
      r9:
        type:  int
        range: [0...2]
      x3:
        type:  choice
        range: {0, 1}
        submodule:
          0:
            x6:
              type:  float
              range: [-1...1]
          1:
            x7:
              type:  float
              range: [-1...1]
\end{lstlisting}

\subsection{OpenML Benchmarks}
\label{openml_spaces}
We define two search spaces with dependencies for two popular machine-learning algorithms (SVM and XGBoost) and evaluate the configurations on 6 most evaluated datasets whose task\_ids are: [10101, 37, 9967, 9946, 10093, 3494]. Furthermore, we compose the two search spaces into a more complex CASH space to further explore the capabilities of our method. In this section, we will give the details of the three search spaces and show the details of the experimental results for each search space on all datasets in Fig.~\ref{fig:openml_benchmark}.

\begin{figure}[t!]
    \centering   
    \includegraphics[width=1\columnwidth]{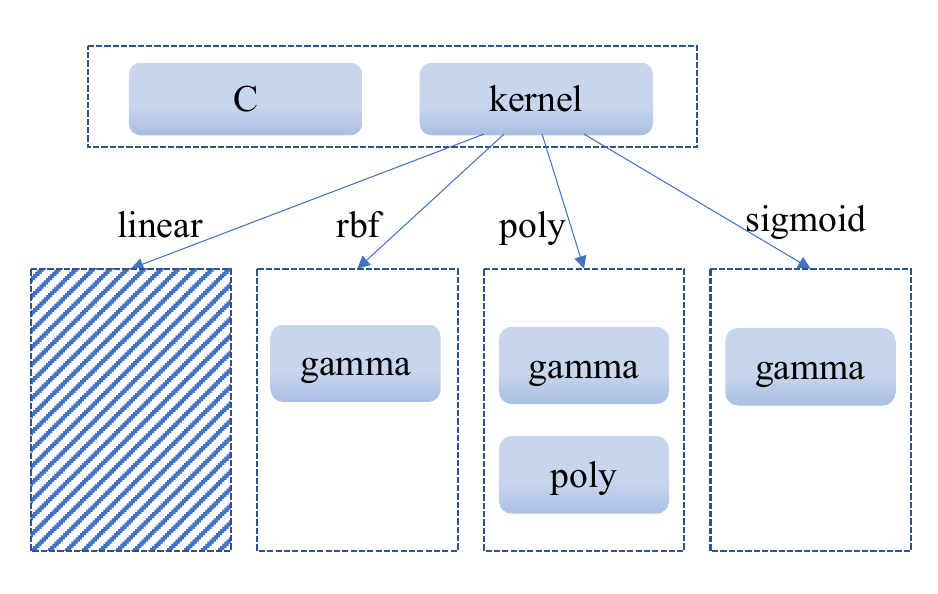}  
\caption{The tree-structured search space of SVM  on the tabular classification tasks. When the kernel is
linear for the SVM model, the shaded box indicates that there are no hyperparameters in this case.}
\label{fig:tree_svm}  
\end{figure}

\subsubsection{SVM Search Space}
\label{svm_space}
The structure of the SVM search space is shown in Listing~\ref{list:svm_space} and Fig.~\ref{fig:tree_svm}. The hyperparameter "kernel" of SVM is a decision hyperparameter --- different "kernel" have different distinct hyperparameters, leading to different subspaces. When the kernel is set to linear, there is no extra hyperparameter and no "submodule" in the YAML file.

\begin{lstlisting}[title={\bf Listing: YAML-style representation of the SVM search space}, label={list:svm_space}, float=tb, caption={YAML of the SVM search space.}, language=yaml] 

C:
  type:  float
  range: [0.001...1000]
kernel:
  type:  choice
  range: {"linear", "poly", "sigmoid", "rbf"}
  submodule:
    poly:
      degree:
        type:  int
        range: [2...6]
      gamma:
        type:  float
        range: [0.001...1000]
    sigmoid:
      gamma:
        type:  float
        range: [0.001...1000]
    rbf:
      gamma:
        type:  float
        range: [0.001...1000]
\end{lstlisting}

\begin{figure}
    \centering   
    \includegraphics[width=1\columnwidth]{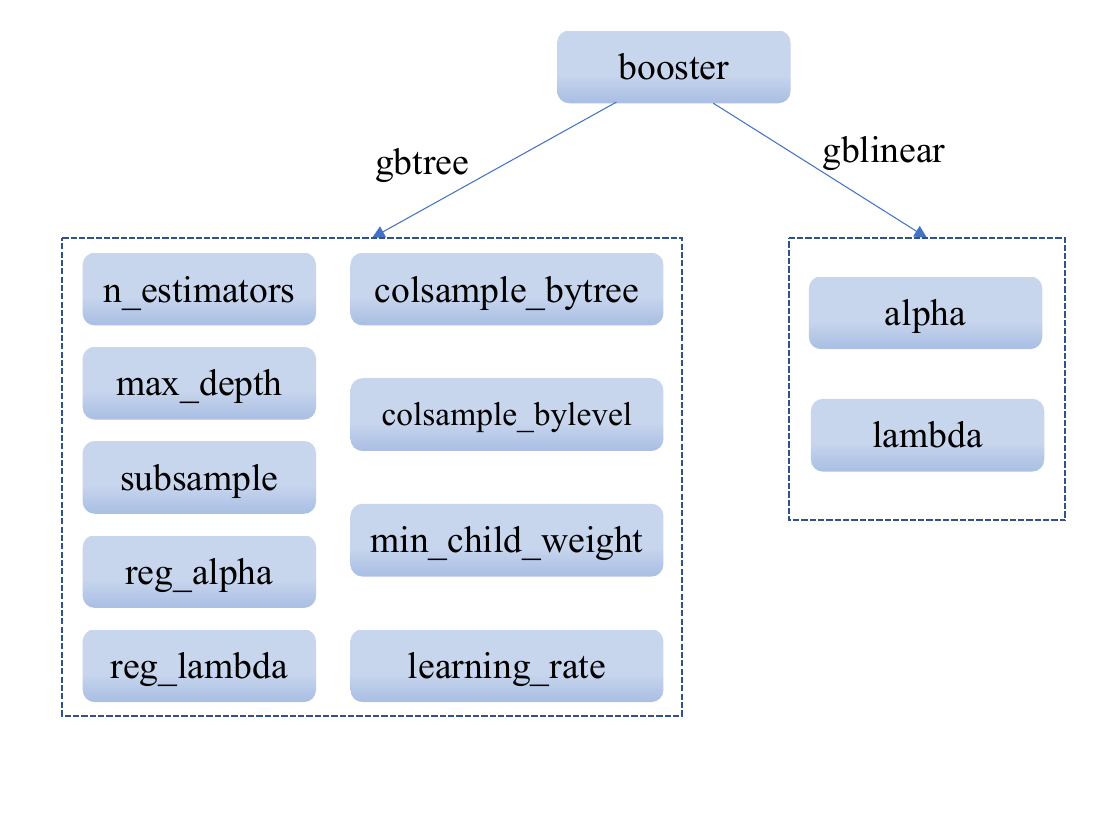}  
\caption{The tree-structured search space of XGBoost on the tabular classification tasks.}
\label{fig:tree_xgb}  
\end{figure}

\subsubsection{XGBoost Search Space}
\label{xgb_space}
The XGBoost search space consists of 10 hyperparameters and is more complex than the SVM search space. Its structure can be seen in Listing~\ref{list:xgb_space} and Fig.~\ref{fig:tree_xgb}. The hyperparameter "booster" of XGBoost is a decision hyperparameter --- different "booster" have different distinct hyperparameters, leading to different subspaces.

\begin{lstlisting}[title={\bf Listing: YAML-style representation of the XGBoost search space}, label={list:xgb_space}, float=t!, caption={YAML of the XGBoost search space.}, language=yaml] 

booster:
  type:  choice
  range: {gbtree, gblinear}
  submodule:
    gbtree:
      n_estimators:
        type:  int
        range: [50...501]
      learning_rate:
        type:  float
        range: [0.001...0.1]
      min_child_weight:
        type:  float
        range: [1...128]
      max_depth:
        type:  int
        range: [1...11]
      subsample:
        type:  float
        range: [0.1...0.999]
      colsample_bytree:
        type:  float
        range: [0.046776...0.998424]
      colsample_bylevel:
        type:  float
        range: [0.046776...0.998424]
      reg_alpha:
        type:  float
        range: [0.001...1000]
      reg_lambda:
        type:  float
        range: [0.001...1000]
    gblinear:
      reg_alpha:
        type:  float
        range: [0.001...1000]
      reg_lambda:
        type:  float
        range: [0.001...1000]
\end{lstlisting}

\begin{figure}[h]
    \centering   
    \includegraphics[width=1\columnwidth]{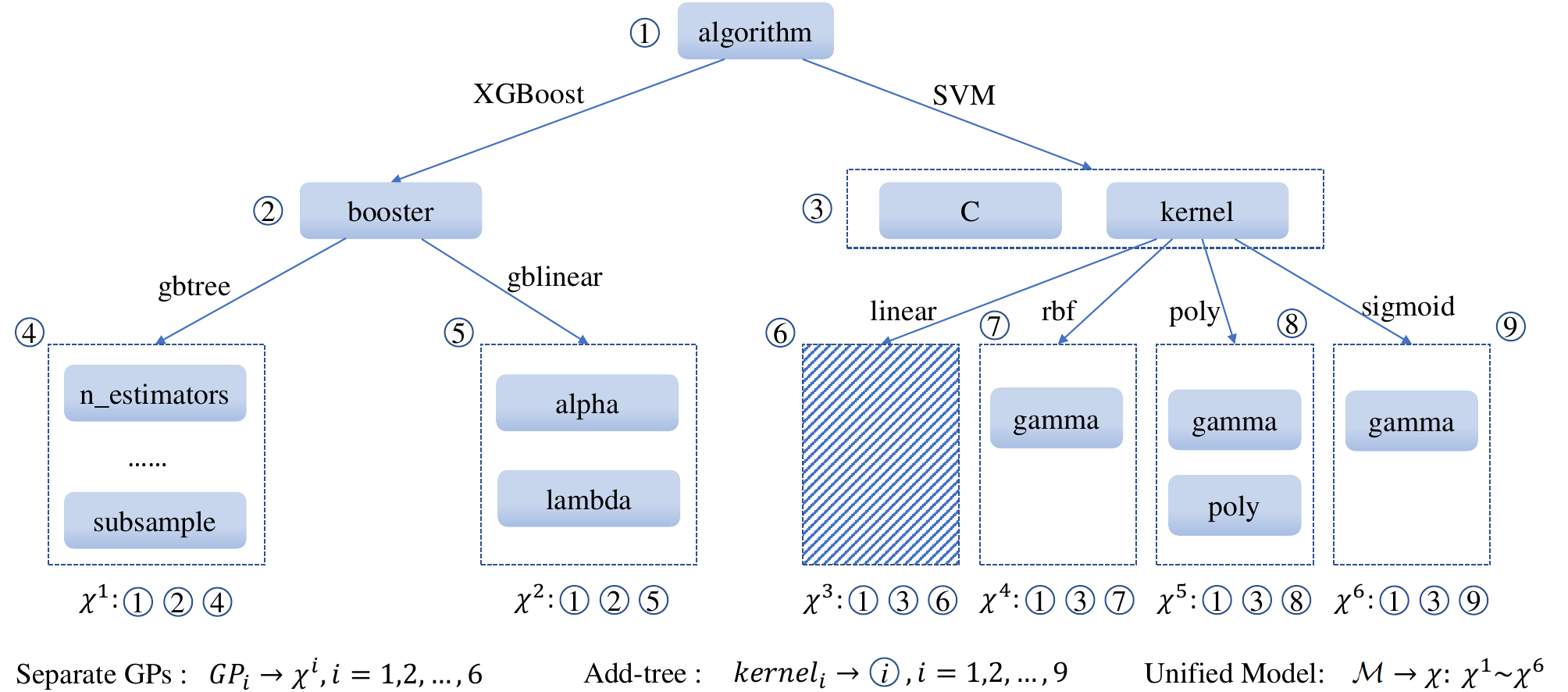}  
\caption{The tree-structured search space on the tabular classification tasks. When the kernel is linear for the SVM model, the shaded box indicates that there are no hyperparameters in this case.}
\label{fig:tree_sp_appendix}  
\end{figure}

\subsubsection{SVM + XGBoost Search Space}
\label{cash_space}
In order to further explore the capabilities of our method, we compose the two search spaces into a more complex CASH space by introducing a meta-level hyperparameter "algorithm" to choose which algorithm will be used to evaluate. The structure of the composed CASH search space is shown in Fig.~\ref{fig:tree_sp_appendix}.

\begin{figure}[h]
    \centering   
    \includegraphics[width=1\columnwidth]{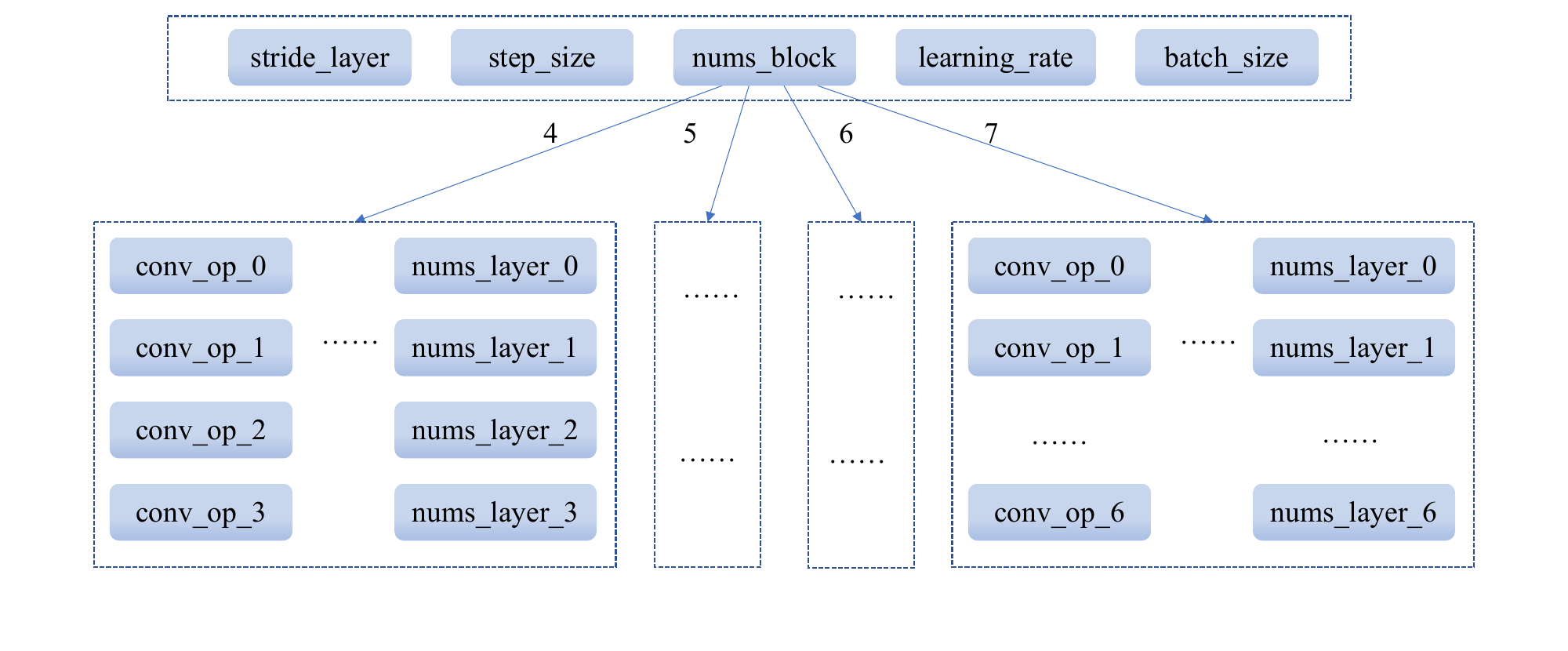}  
\caption{The tree-structured search space of the NAS task evaluated on CIFAR-10 dataset.}
\label{fig:tree_mnas}  
\end{figure}

\begin{figure}
    \centering   
    \includegraphics[width=1\columnwidth]{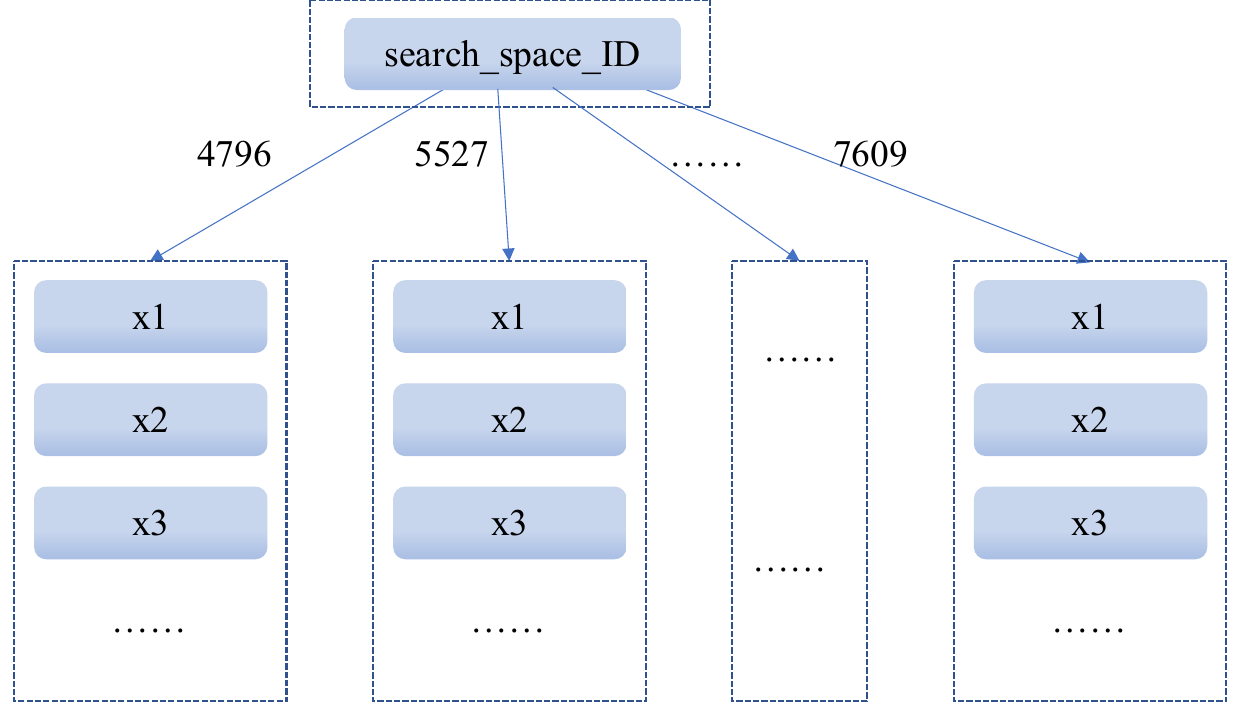}  
\caption{The tree-structured search space of the HPOB task.}
\label{fig:tree_hpob}  
\end{figure}

\begin{figure*}[tb]
    \centering
  \subfigure[standard convolution]{
    \label{fig:nas_std}
    \includegraphics[width=0.46\textwidth]
    {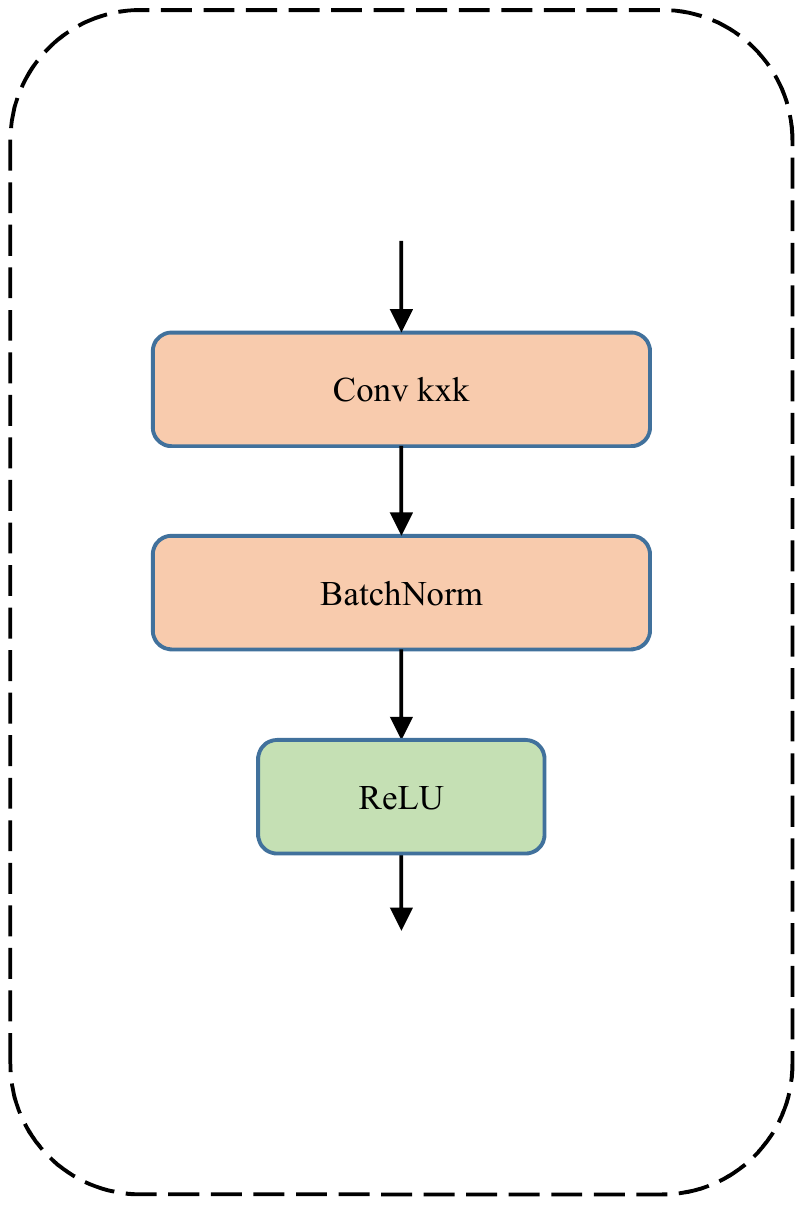}
    }
  \subfigure[depthwise separable convolution]{
    \label{fig:nas_dep}
    \includegraphics[width=0.47\textwidth]{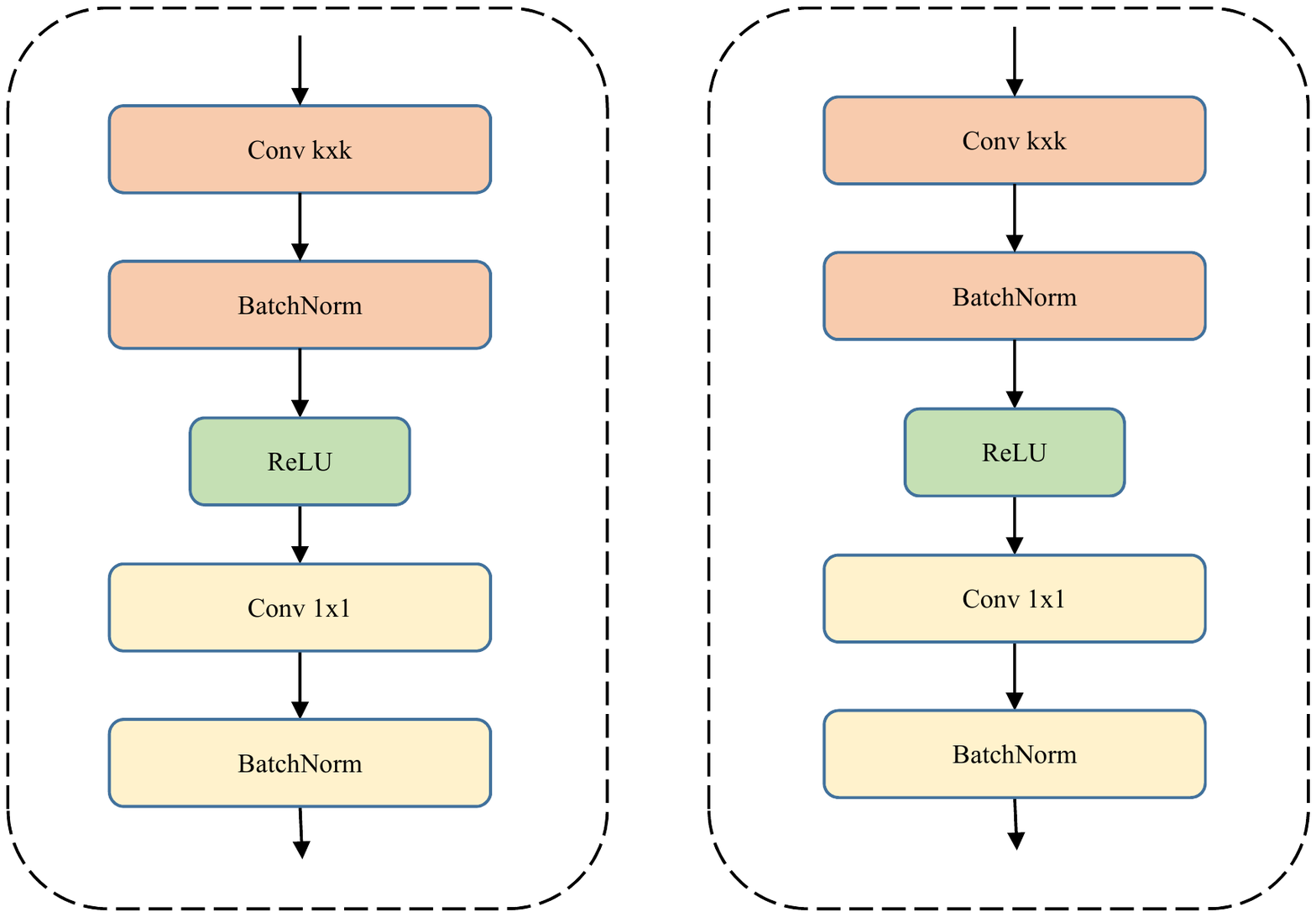}  
    }
    
  \subfigure[inverted residual layer]{
    \label{fig:nas_inv}
    \includegraphics[width=1\columnwidth]{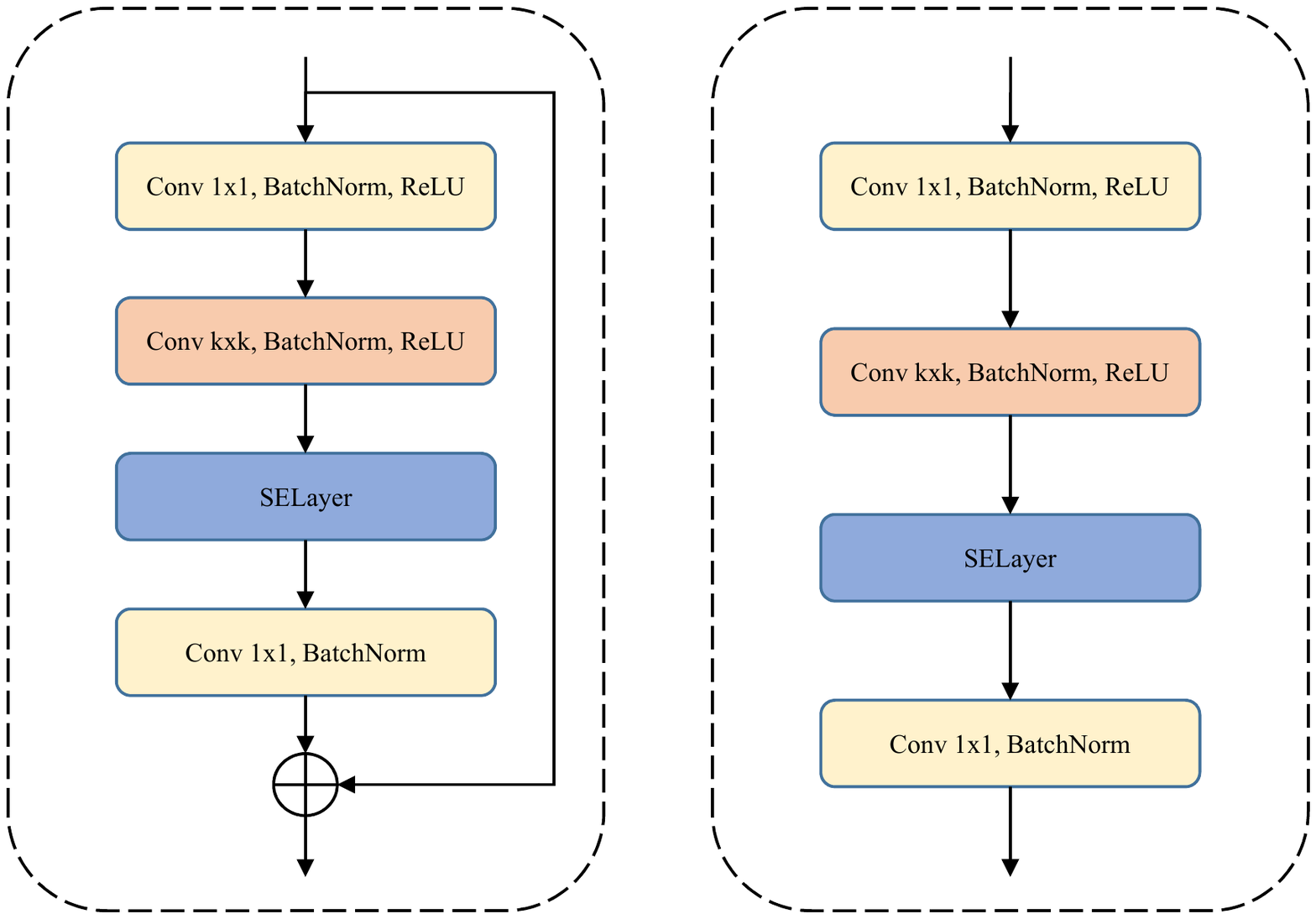}  
    }
  \subfigure[residual layer]{
    \label{fig:nas_res}
    \includegraphics[width=1\columnwidth]{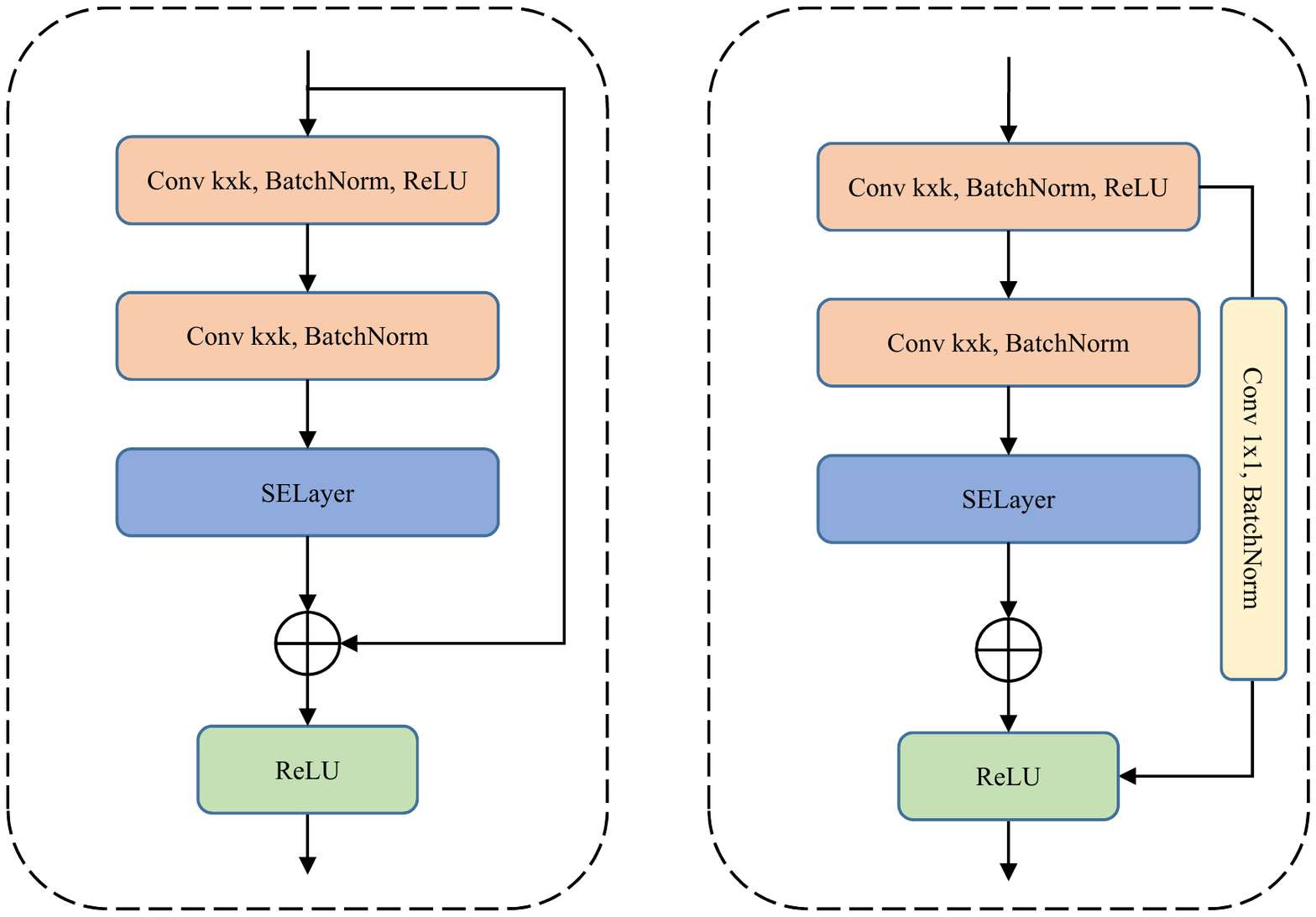}  
    }
\caption{Searched architectures for NAS task.}
\label{fig:mnas_arch}
\end{figure*}

\subsection{Neural Architecture Search}
\label{nas_space}
Following~\citet{MNAS}, we define a factorized hierarchical search space to find the best network architecture and its training configurations. The search space consists of two aspects: 1) Neural network architectures. 2) Hyperparameters of the optimizer used for training the neural networks.

As shown in Fig.~\ref{fig:mnas_arch}, for the network architectures, we group the network layers into a number of provisioned skeletons, called blocks, based on some solid works~\citep{mobile_net,mobile_net2,MNAS,efficient_net} in computer vision. Each block contains various repeated identical layers, except striding. Only the first layer has stride 2 if the block needs to downsample, while all other layers have stride 1. We use the hyperparameter "stride\_layer" to control this operation. For each block, we search for the types of stacked convolution operations and connections for a single layer and the number of layers "nums\_layer"($N$), and then for one layer $i$ is repeated
$N_i$ times (e.g., Layer 4-1 to 4-N\_4 are the same, where N\_4 represents the number of repeated layers in the 4th block). Now we describe the details of each hyperparameter:
\begin{enumerate}
\item $nums\_block$. The number of blocks.
\item $conv\_op$. The convolution operation type for a single layer of each block. In our settings, following~\citet{MNAS}, there are 4 provisioned types available, represented by codes 0, 1, 2, and 3 respectively. 
1) The first is the standard convolution layer~\citep{vgg}, which consists of a 2D convolution operation with a kernel size of $(kernel\_size \times kernel\_size)$, a batch normalization operation and a ReLU activation function. 
2) The second type is the depthwise separable convolution layer~\citep{mobile_net}. It has the same function as the standard convolution layer but is more efficient, which is a form of factorized convolutions with a standard convolution into a depthwise
convolution and a 1×1 convolution called a pointwise convolution.  
3) The next one is the inverted residual layer~\citep{mobile_net2}, where each layer contains an input followed
by two bottlenecks and two expansion layers between them. 4) The last type is the ResNet layer commonly used in computer vision tasks~\citep{resnet}.
\item $kernel\_size$. The size of the convolution kernel in one convolution block. 
\item $nums\_layer$. The number of layers in each block.
\item $expend\_ration$. The ratio for expending, if using the inverted residue block~\citep{mobile_net2}.
\item $seratio$. The ratio of squeezing and expending if containing such structure.
\item $nums\_channel$. The number of channels for each block.
\item $stride\_layer$. The number of strides for each block is represented in binary.
\end{enumerate}

The optimization hyperparameters
The details of each hyperparameter are as follows:
\begin{enumerate}
\item $learning\_rate$. The learning rate determines the speed of the network's training and convergence.
\item $step\_size$. Size of the change in the parameter when the optimizer updates the parameter. 
\item $batch\_size$. Batch size determines how many data points will be used for training in each iteration.
\end{enumerate}

The structure of the search space of the NAS task is shown in Listing~\ref{list:nas_space} and Fig.~\ref{fig:tree_mnas}.

\begin{lstlisting}[title={\bf Listing: YAML-style representation of the NAS search space}, label={list:nas_space}, float=tb, caption={YAML of the NAS search space.}, language=yaml] 

# hyperparameters of the network architecture
nums_block:
  type: int
  range:
  - 4...8
  submodule:
    conv_op:
      type: choice
      range: {0, 1, 2, 3}
    expand_ratio:
      type:  int
      range:  [5...7]
    seratio:
      type: choice
      range: {0, 8, 16}
    kernel_size:
      type: choice
      range: {3, 5}
    nums_layer:
      type: choice
      range: {0, 1, 2}
    nums_channel:
      type: choice
      range: {1, 1.25, 1.3}
stride_layer:
  type: choice
  range: {43, 44}

# hyperparameters for optimization
learning_rate:
  type: float
  range: [0.07...0.15]
step_size:
  type: int
  range: [70...90]
batch_size:
  type: powerint2
  range: [5...8]
\end{lstlisting}

\subsection{Meta-Learning on HPO-B-v3 Benchmark}
HPO-B-v3~\citep{HPOB} is a large-scale hyperparameter optimization benchmark, which contains a collection of 935 black-box tasks for 16 hyperparameter search spaces evaluated on 101 datasets. We set the ID of a search space as the father node of its hyperparameters, resulting in a tree-structured search space shown in Fig.~\ref{fig:tree_hpob}. We meta-train our model on all training data points of the 16 search spaces and fine-tune it on the test tasks to get the final performance. We set the learning rate of the attention-based encoder and the GP to 1e-5 and 1e-3, respectively. In the meta-training stage, we train the attention-based DKGP for 300 epochs on all training data points. Then, we use the trained weights as the initialization on new tasks. We show our performance on each search space in Fig.~\ref{fig:hpob_regret} and Fig.~\ref{fig:hpob_ranking}.

\begin{figure*}[h]
    \centering  \includegraphics[width=2.0\columnwidth]{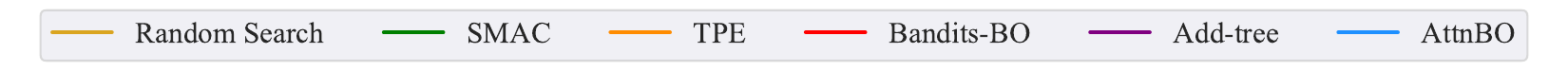}

  \subfigure[SVM 10101]{
    \includegraphics[width=0.55\columnwidth]{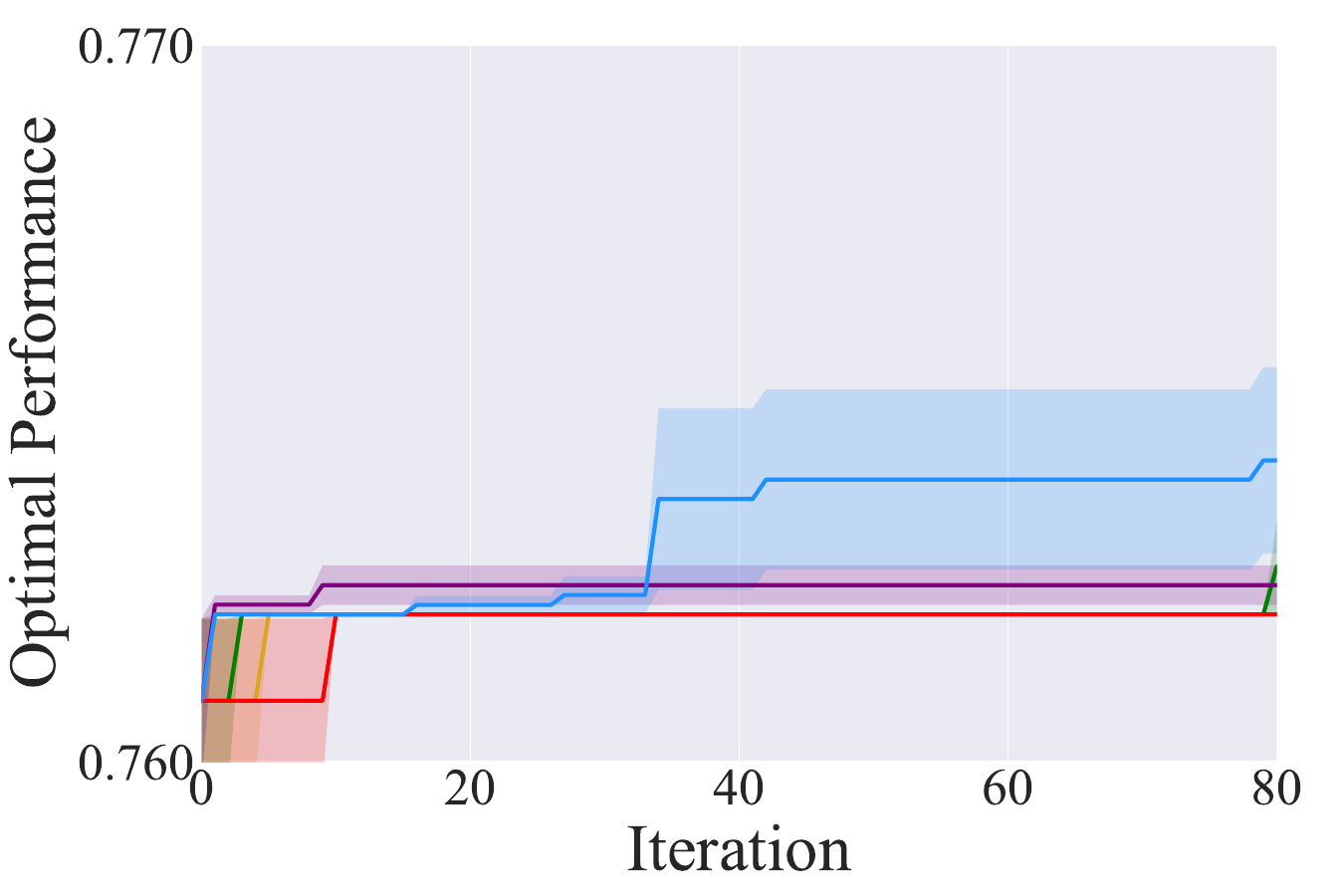}  
}
  \subfigure[SVM 37]{
    \includegraphics[width=0.55\columnwidth]{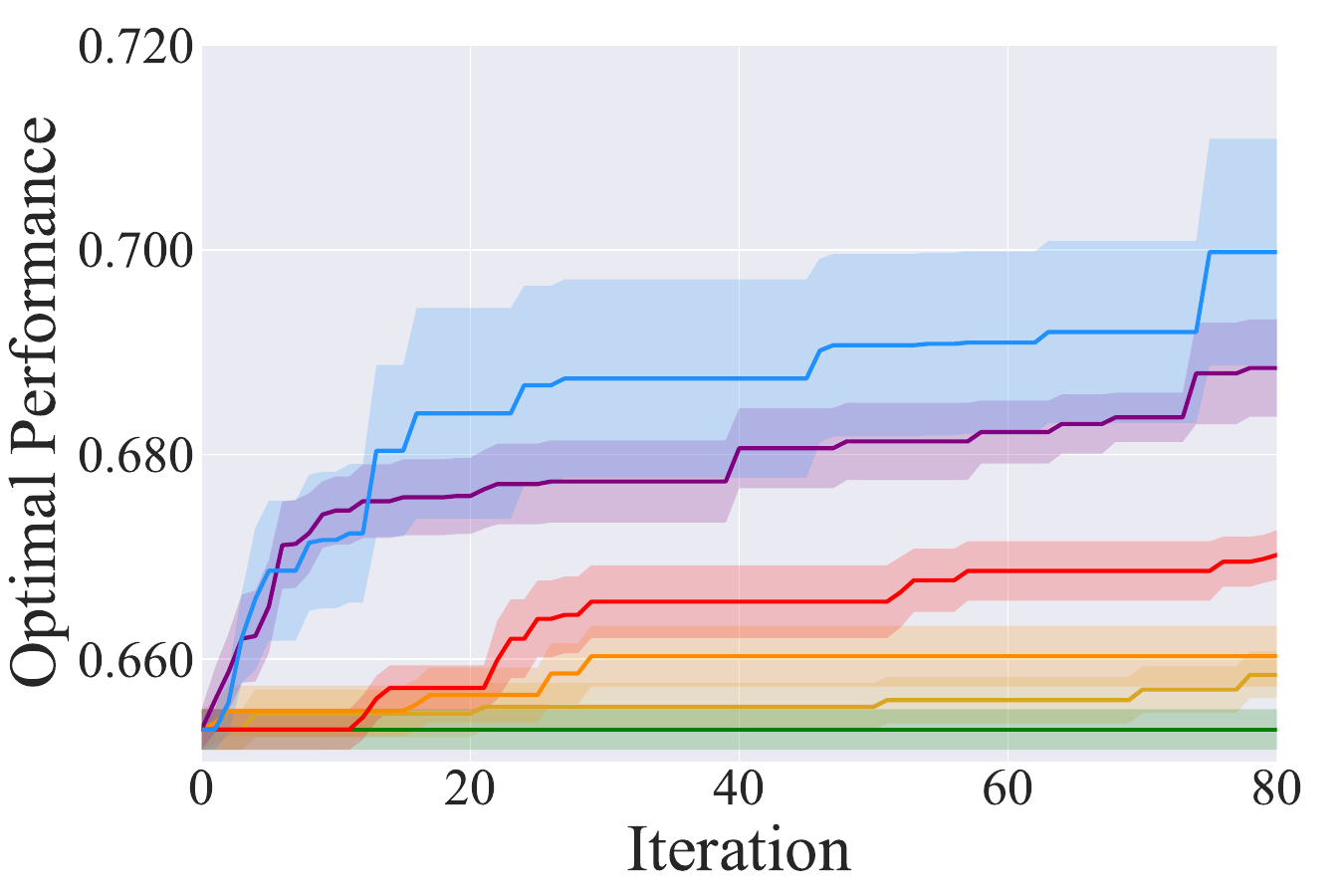}  
    }
  \subfigure[SVM 9967]{
    \includegraphics[width=0.55\columnwidth]{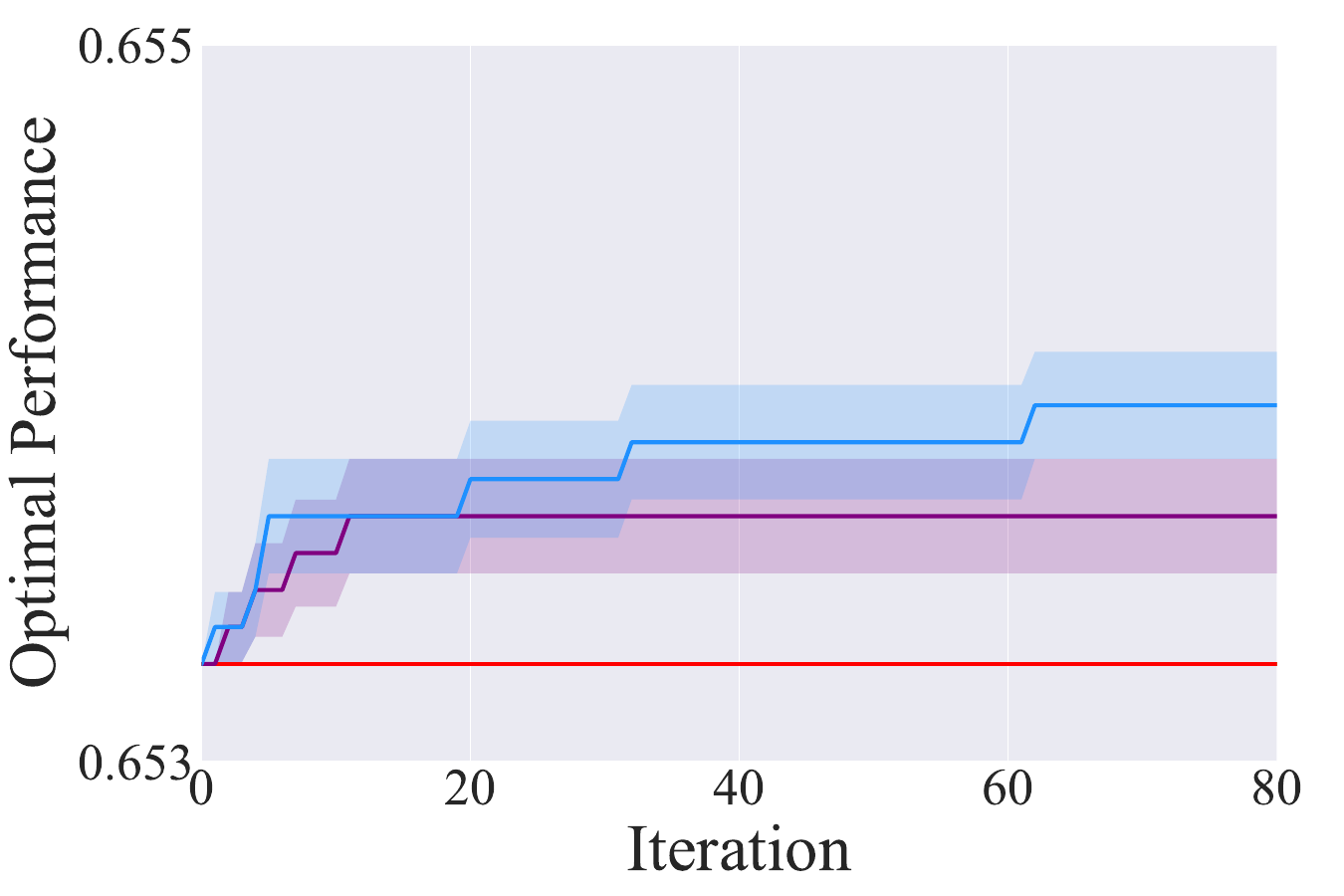}  
    }
   \subfigure[SVM 10093]{
    \includegraphics[width=0.55\columnwidth]{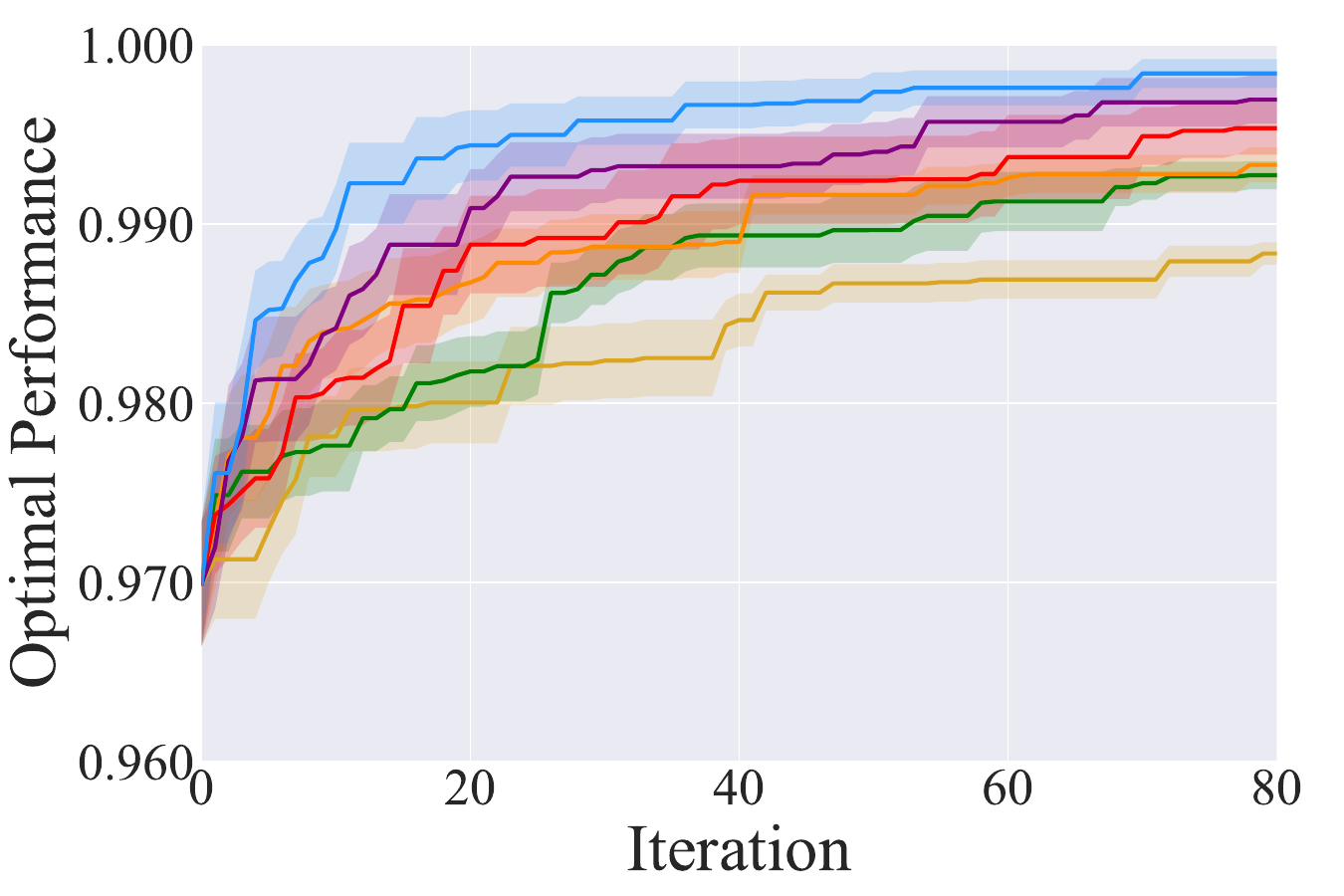}  
    }
  \subfigure[SVM 9946]{
    \includegraphics[width=0.55\columnwidth]{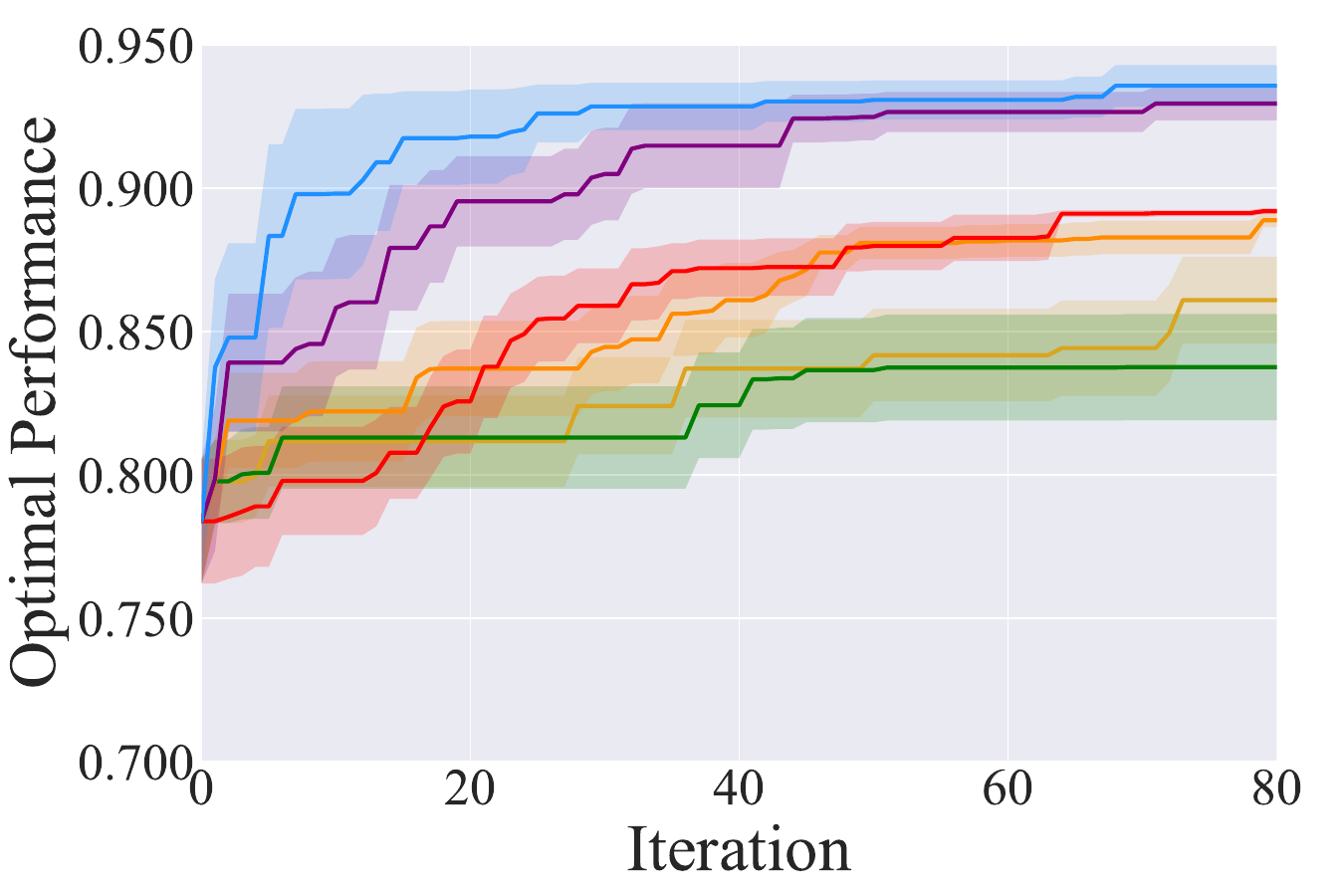}  
    }
  \subfigure[SVM 3494]{
    \includegraphics[width=0.55\columnwidth]{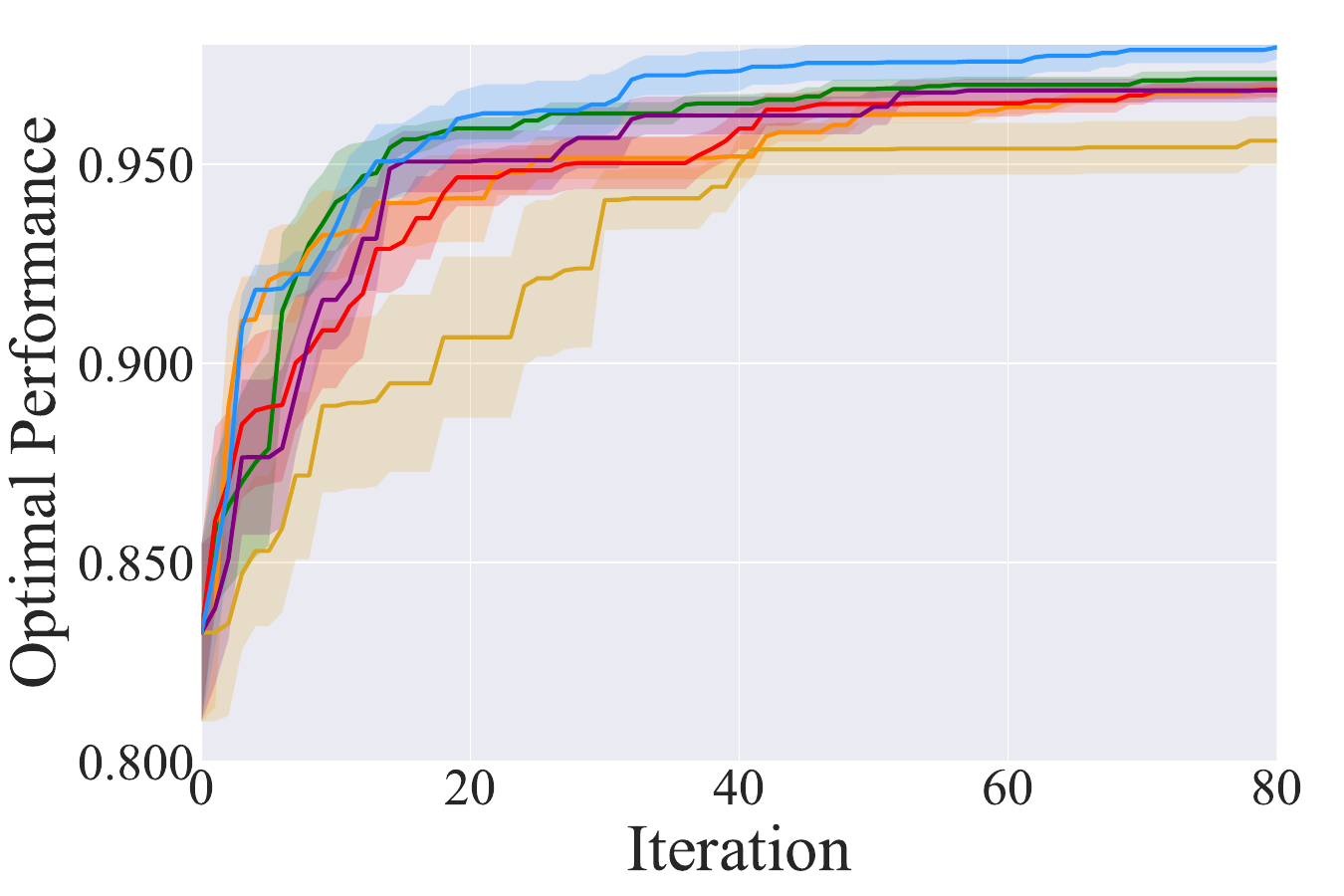}  
    }

  \subfigure[XGBoost 10101]{
    \includegraphics[width=0.55\columnwidth]{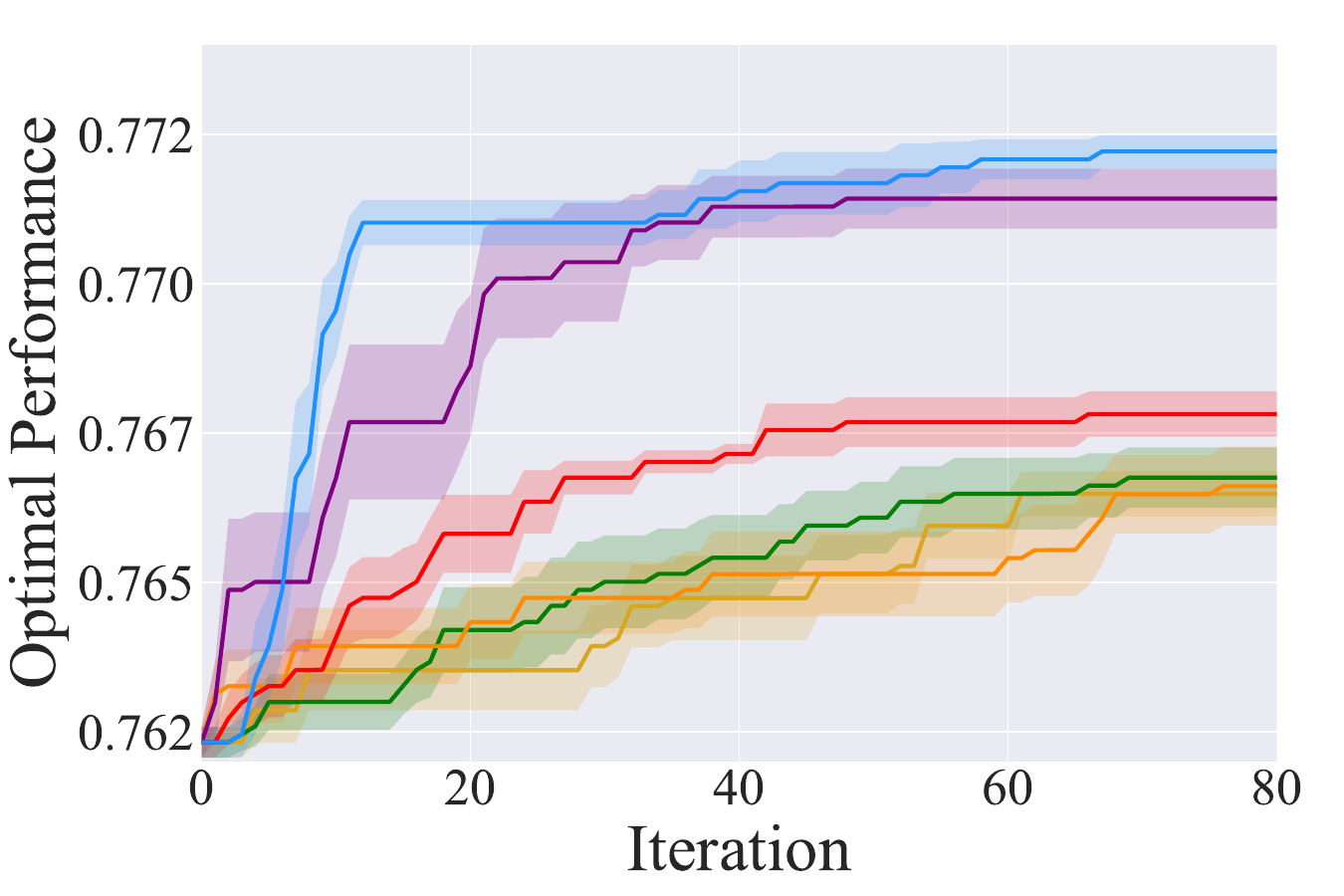}  
    }
  \subfigure[XGBoost 37]{
    \includegraphics[width=0.55\columnwidth]{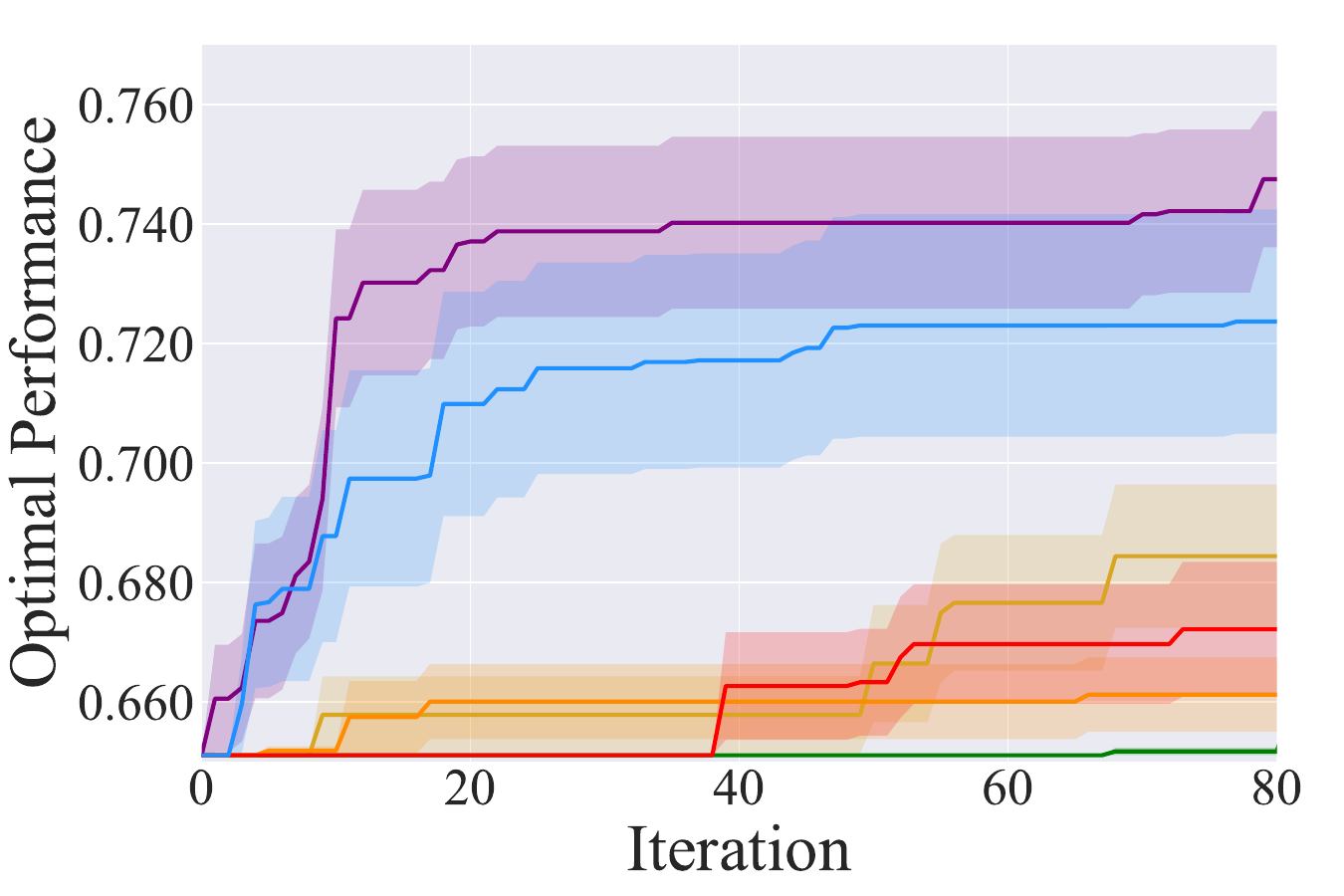}  
    }
  \subfigure[XGBoost 9967]{
    \includegraphics[width=0.55\columnwidth]{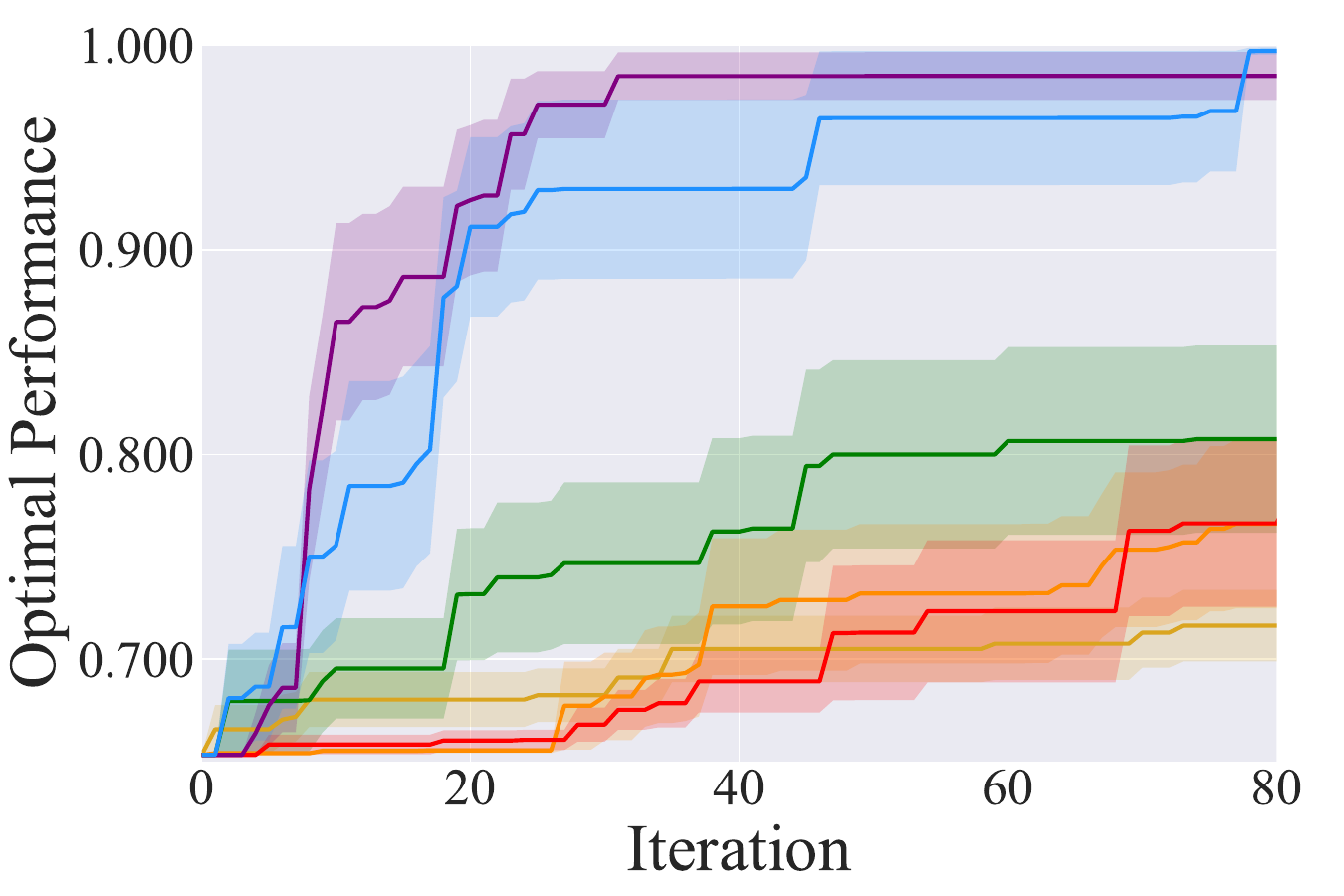}  
    }
   \subfigure[XGBoost 10093]{
    \includegraphics[width=0.55\columnwidth]{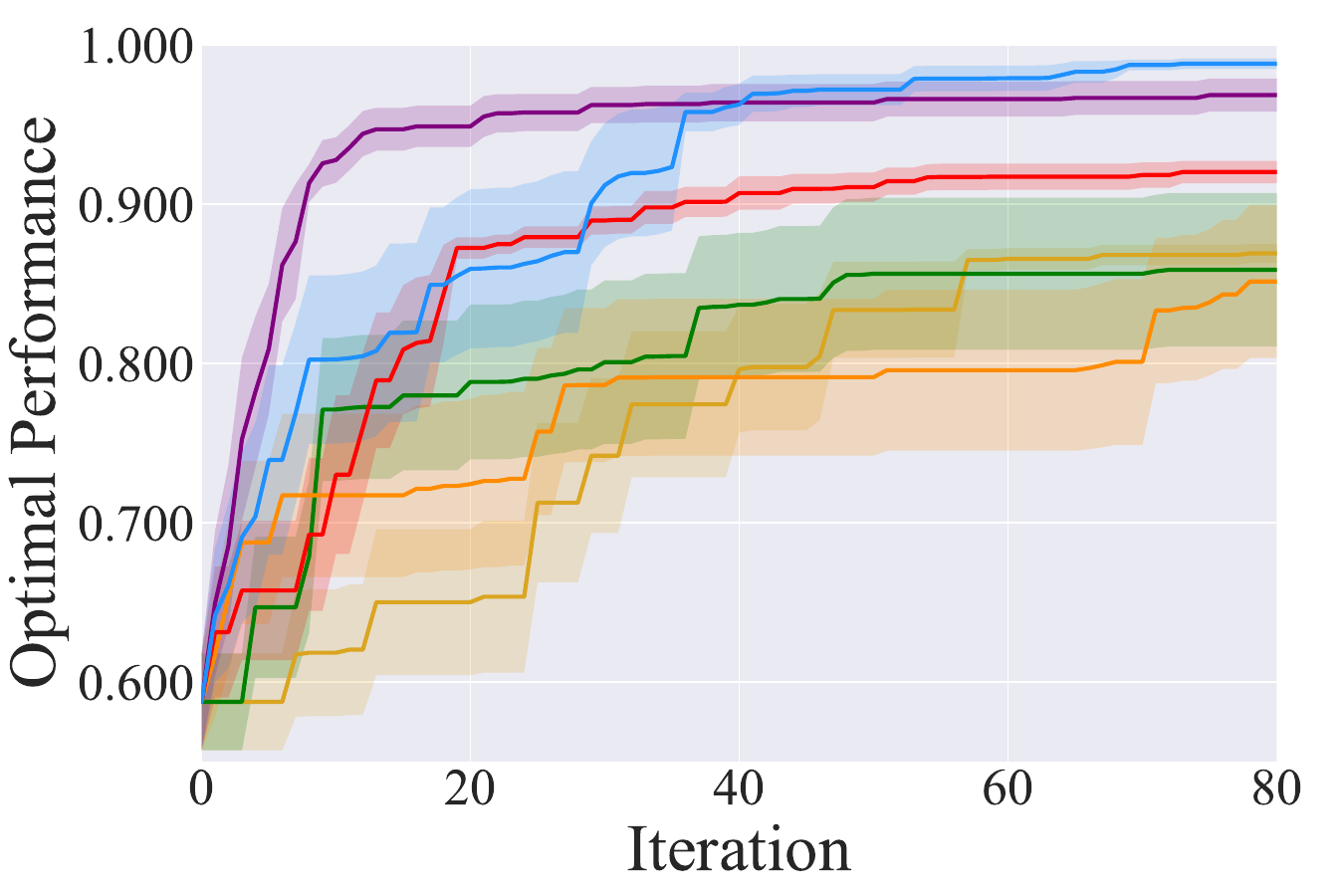}  
    }
  \subfigure[XGBoost 9946]{
    \includegraphics[width=0.55\columnwidth]{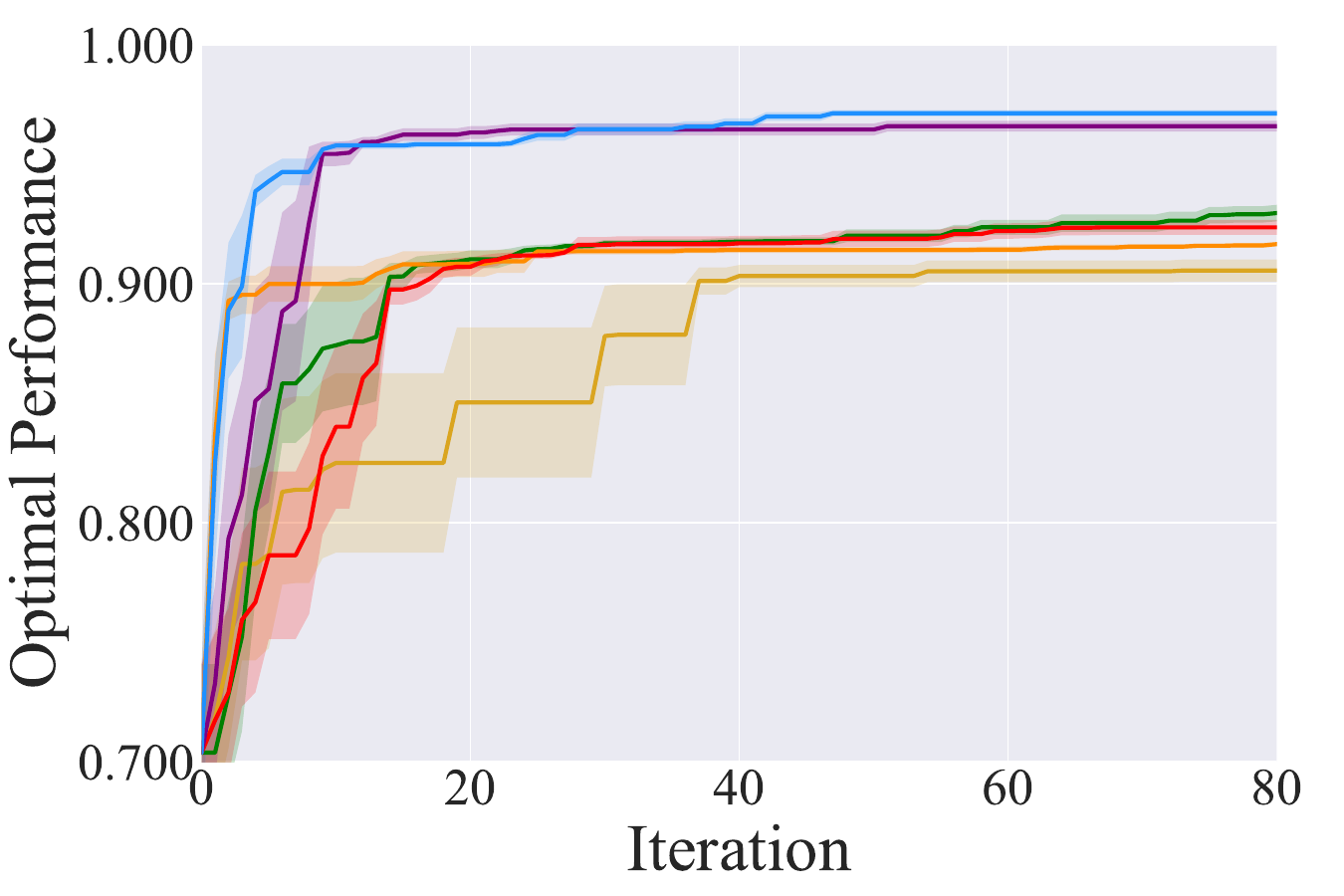}  
    }
  \subfigure[XGBoost 3494]{
    \includegraphics[width=0.55\columnwidth]{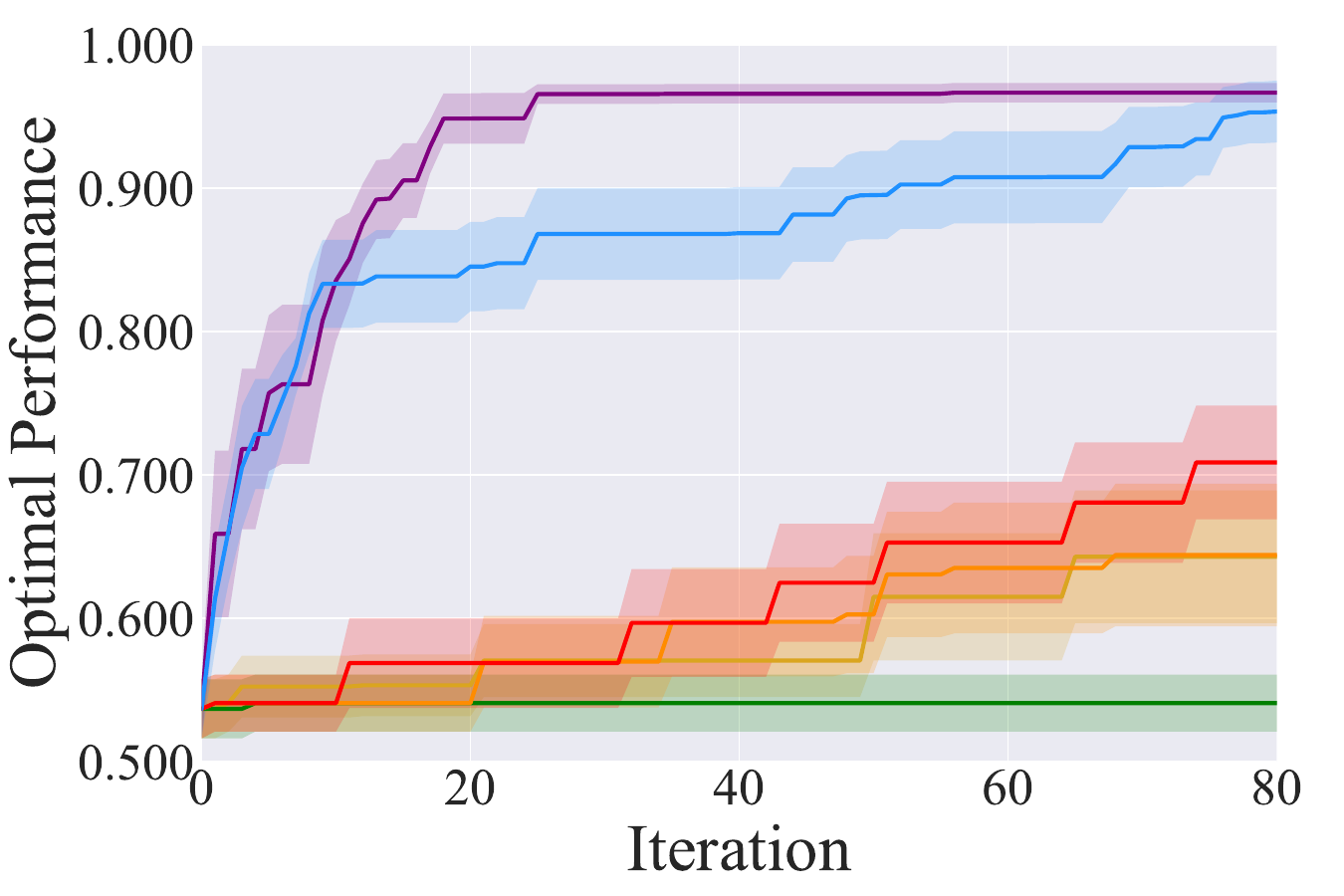}  
    }
  \subfigure[SVM+XGBoost 10101]{
    \includegraphics[width=0.55\columnwidth]{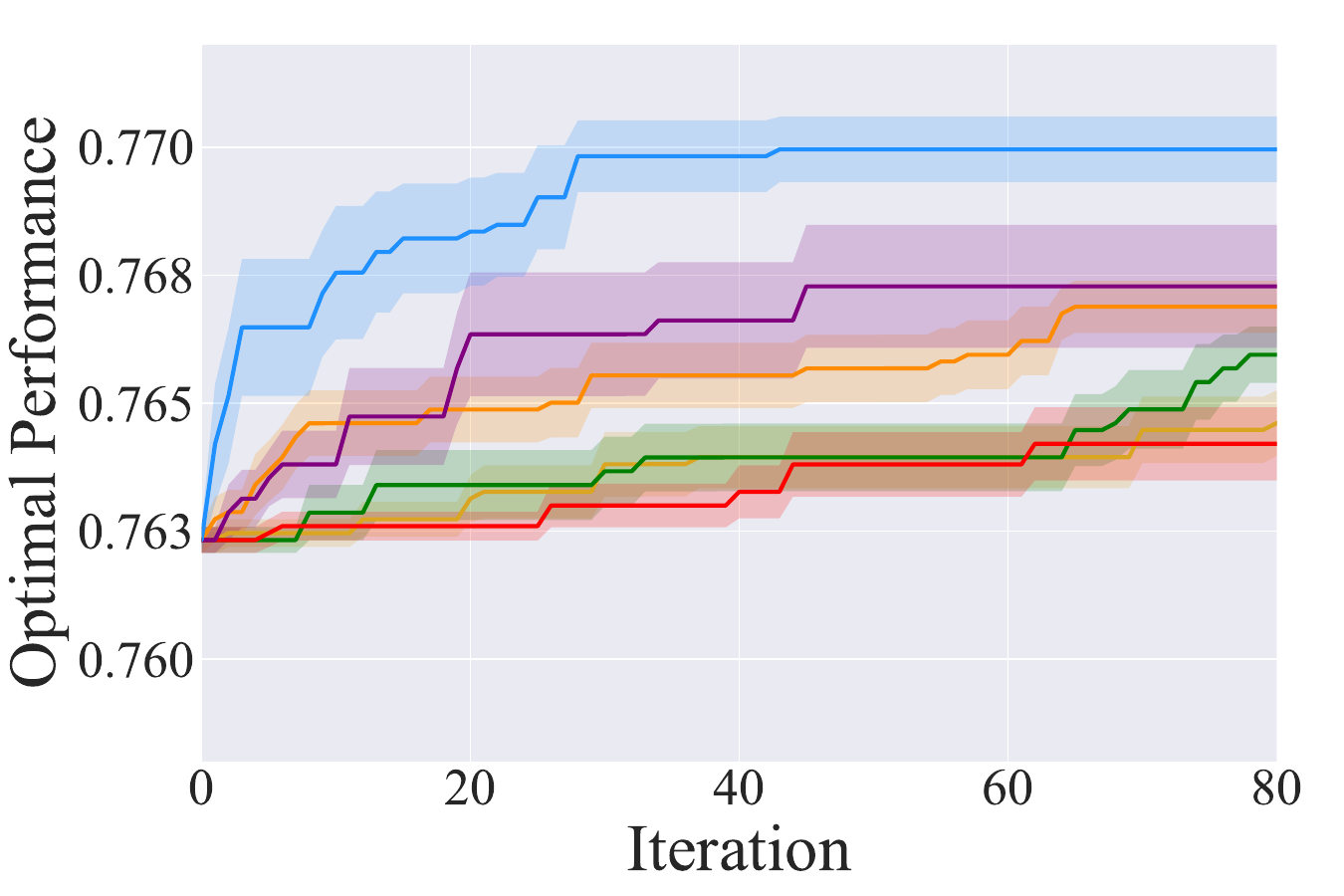}  
    }
  \subfigure[SVM+XGBoost 37]{
    \includegraphics[width=0.55\columnwidth]{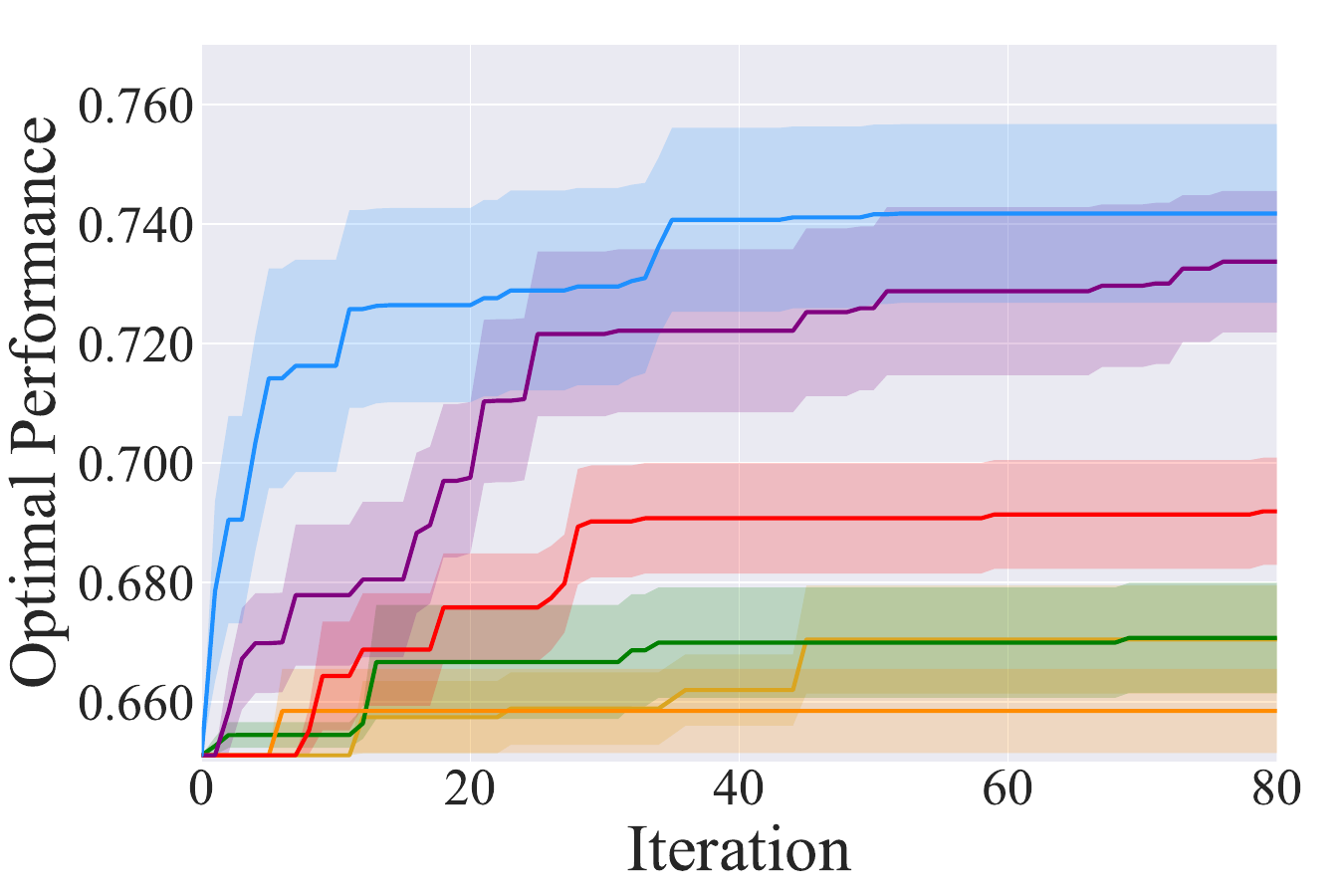}  
    }
  \subfigure[SVM+XGBoost 9967]{
    \includegraphics[width=0.55\columnwidth]{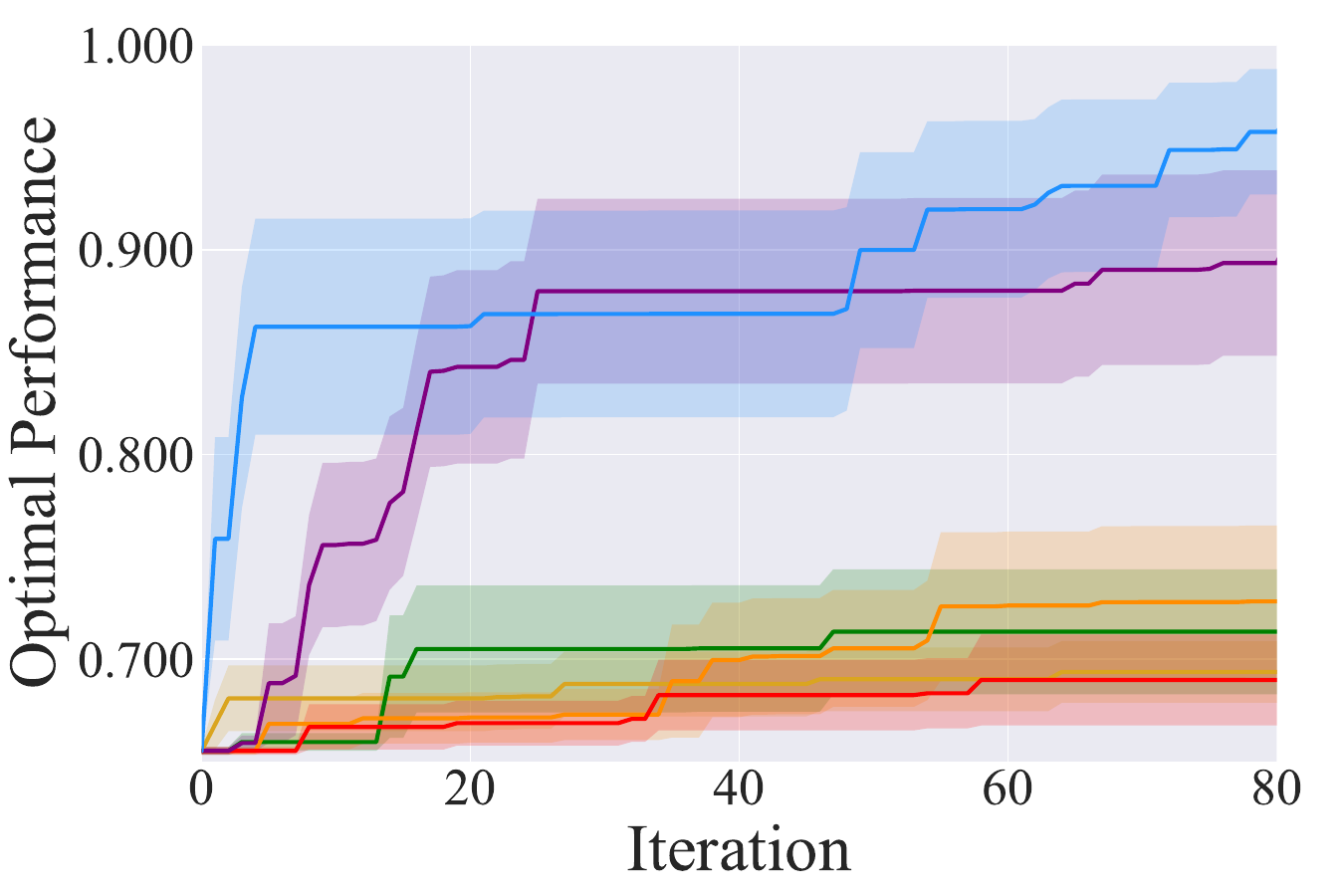}  
    }
   \subfigure[SVM+XGBoost 10093]{
    \includegraphics[width=0.55\columnwidth]{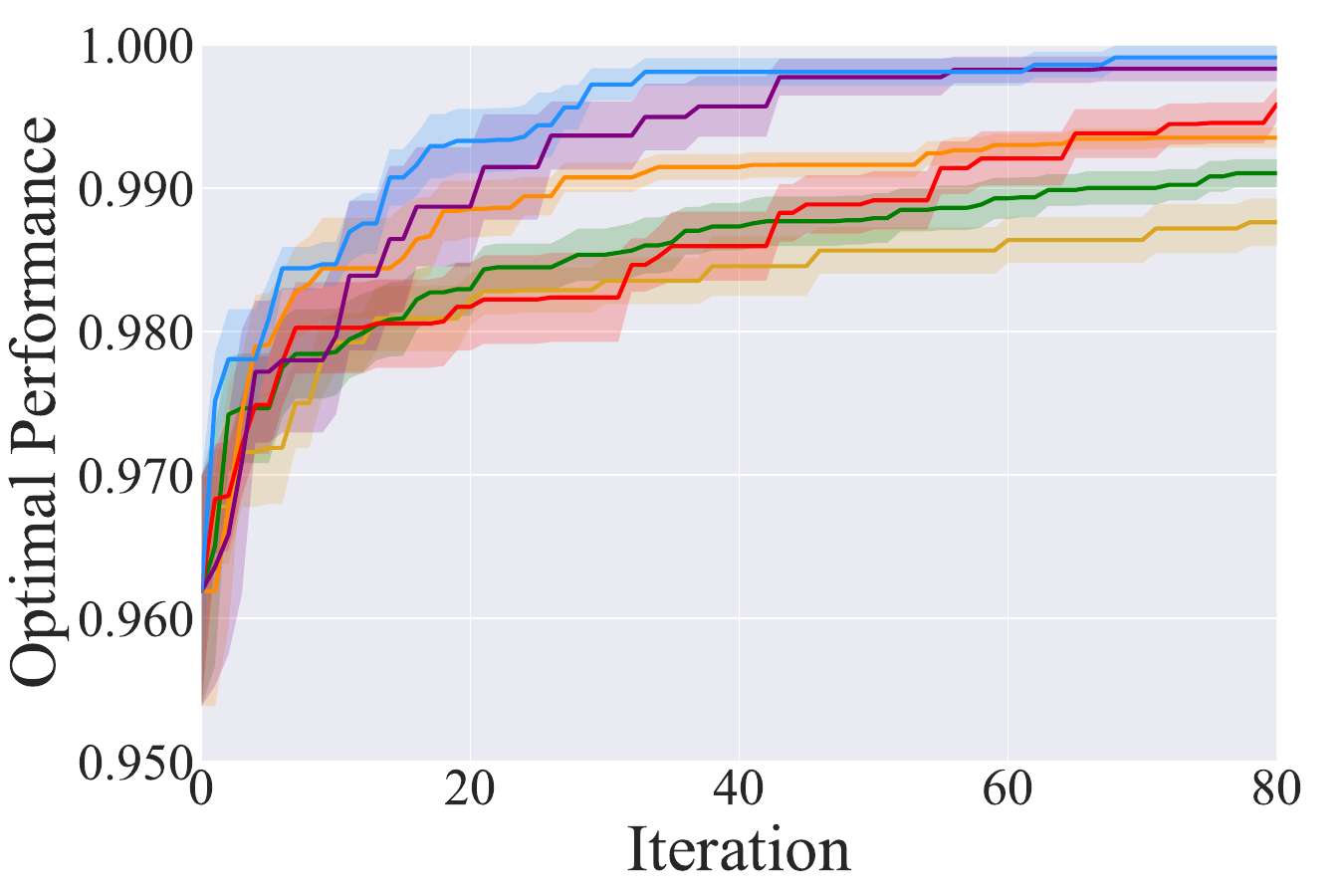}  
    }
  \subfigure[SVM+XGBoost 9946]{
    \includegraphics[width=0.55\columnwidth]{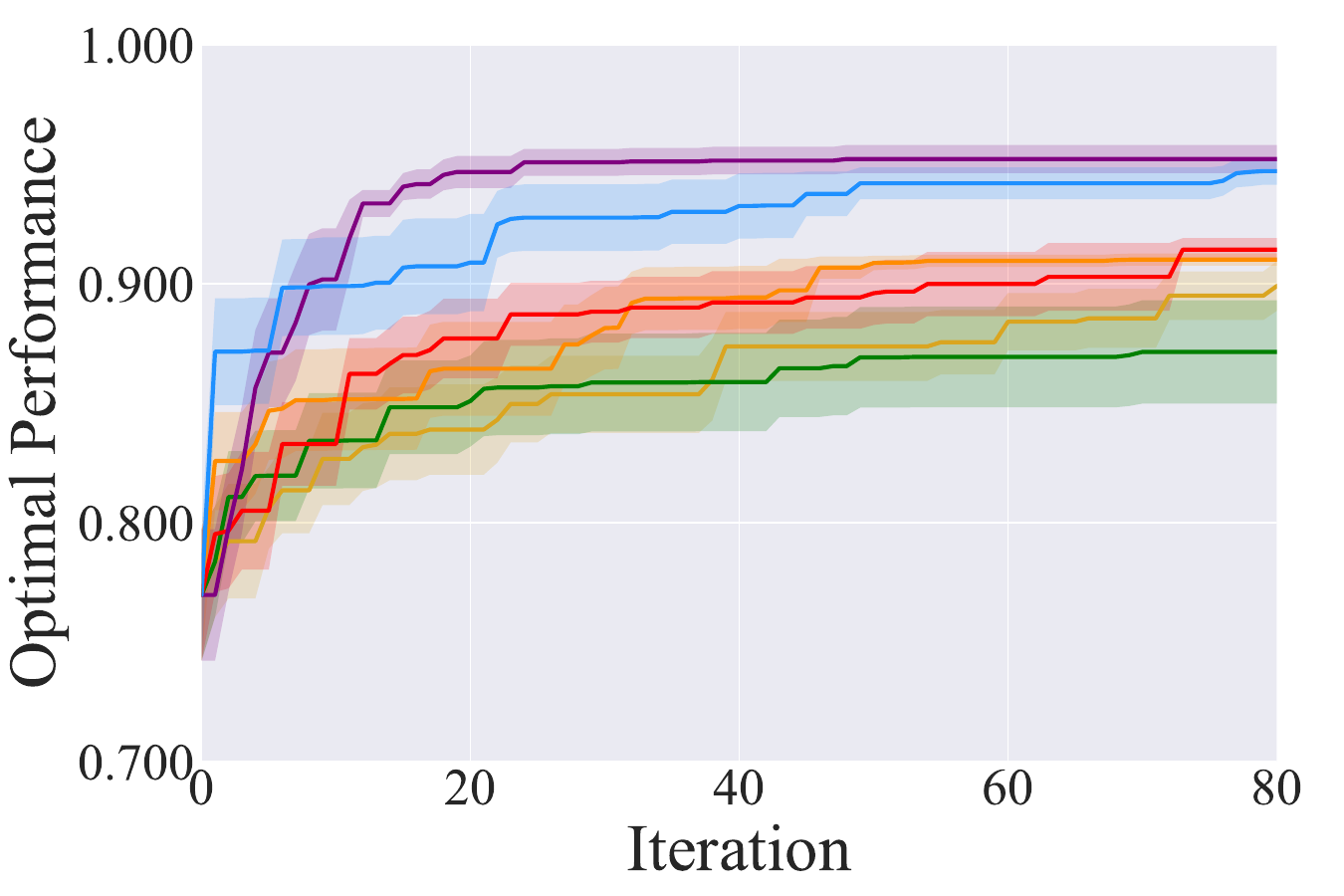}  
    }
  \subfigure[SVM+XGBoost 3494]{
    \includegraphics[width=0.55\columnwidth]{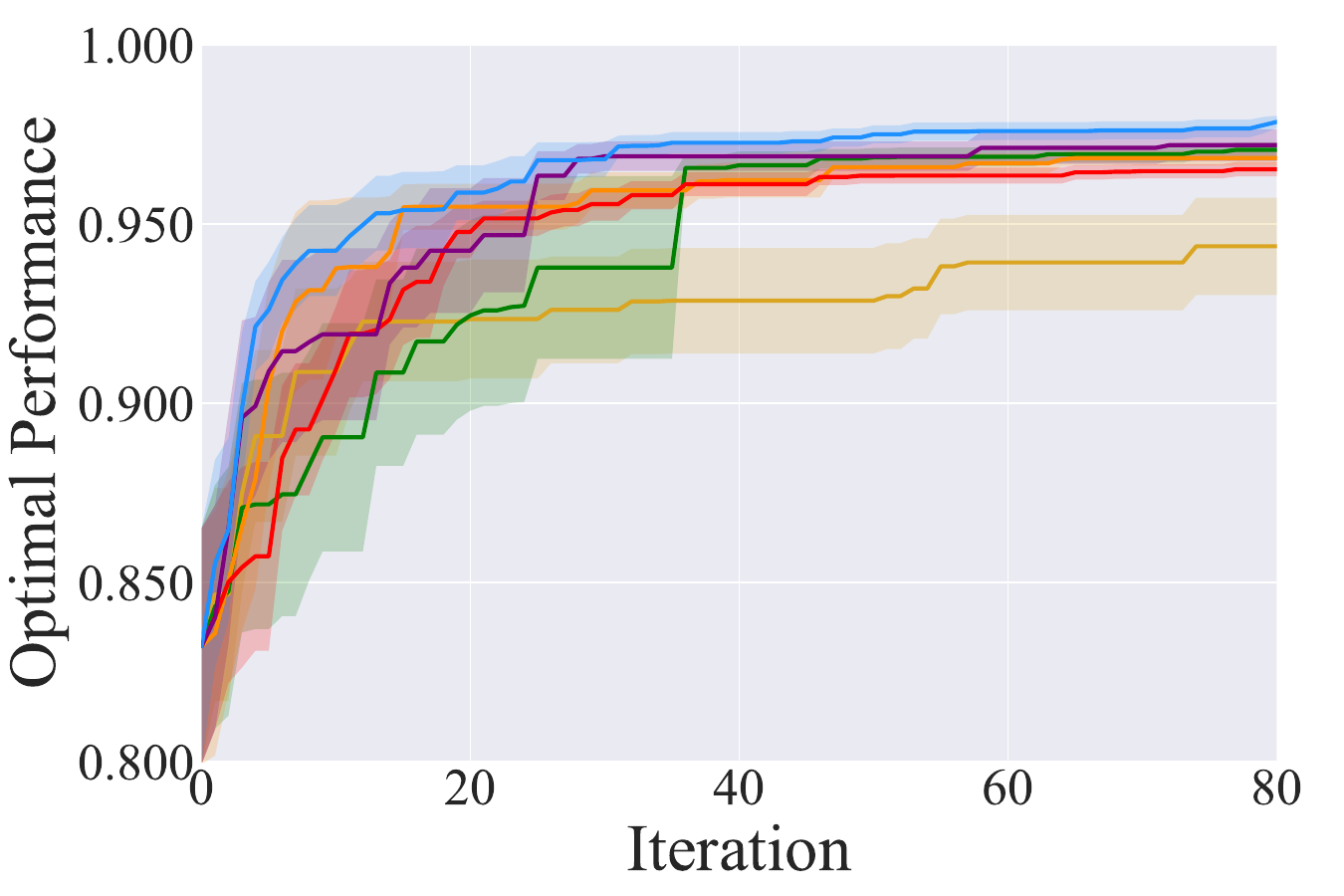}  
    }
  \caption{Performance of various black-box optimization methods on three machine-learning benchmarks evaluated on real-world OpenML datasets.}
\label{fig:openml_benchmark}
\end{figure*}

\begin{figure*}[h]
    \centering  
\includegraphics[width=1.0\textwidth]
    {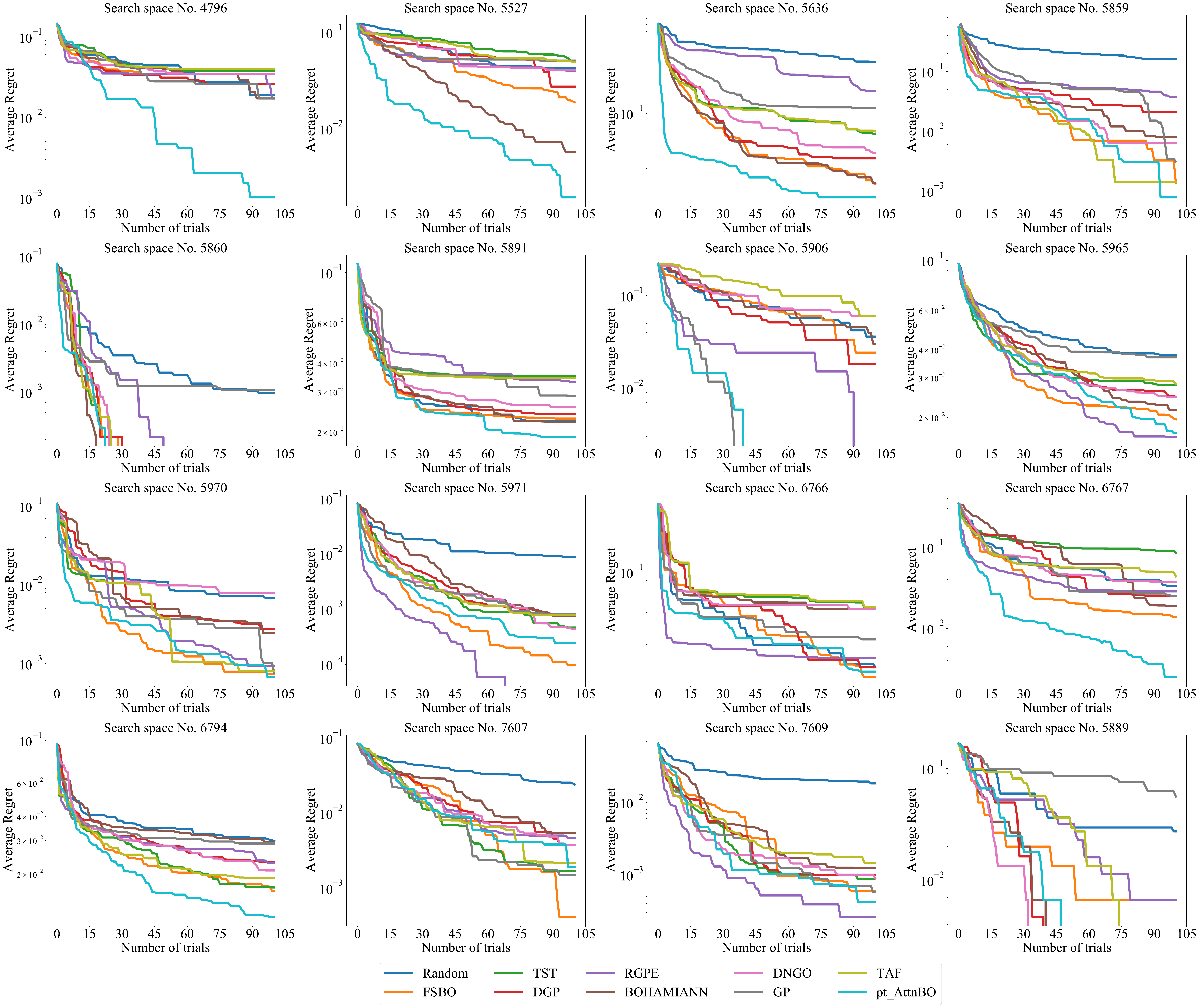}
 
  \caption{Average regrets of various methods on HPO-B-v3 benchmark.}
\label{fig:hpob_regret}
\end{figure*}

\begin{figure*}[h]
    \centering  
\includegraphics[width=1.0\textwidth]
    {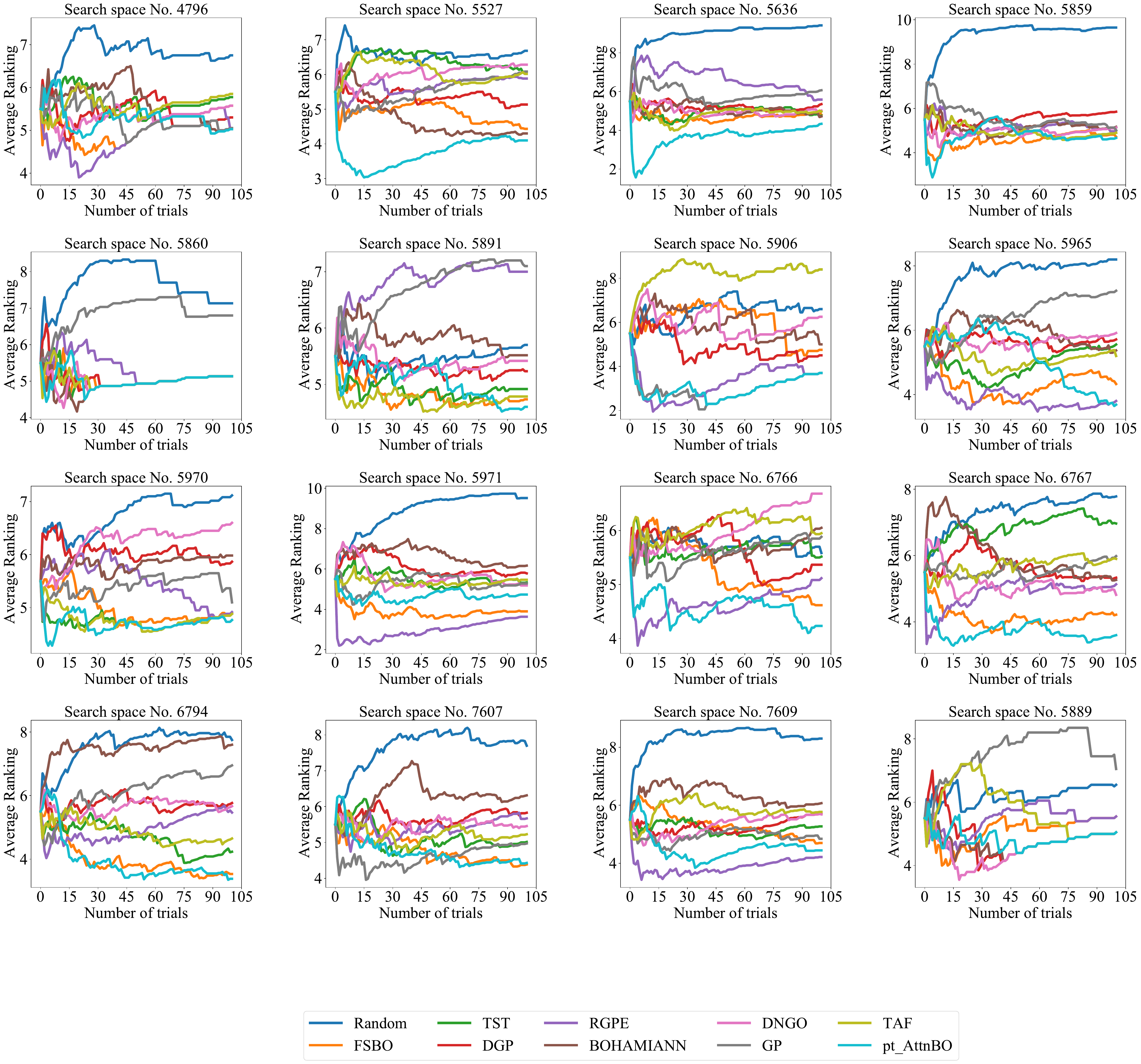}
 
  \caption{Average rankings of various methods on HPO-B-v3 benchmark.}
\label{fig:hpob_ranking}
\end{figure*}

\end{document}